\documentclass[5p]{elsarticle}
\usepackage{hyperref}
\usepackage{times}
\usepackage{epsfig}
\usepackage{graphicx}
\usepackage{amsmath}
\usepackage{amssymb}
\usepackage{stmaryrd}
\usepackage{eucal}
\usepackage{bm}
\usepackage{epsfig}
\usepackage{subfigure}
\usepackage{caption}
\usepackage{xspace}
\usepackage{enumerate}
\usepackage{multirow}

\journal{Neural Networks}

\begin{document}

\begin{frontmatter}
\title{Block-term Tensor Neural Networks}

\author[address1]{Jinmian Ye \corref{equalcontribution}}
\author[address2]{Guangxi Li \corref{equalcontribution}}
\author[address1]{Di Chen}
\author[address3]{Haiqin Yang}
\author[address4]{Shandian Zhe}
\author[address5,address6,address1]{Zenglin Xu\corref{mycorrespondingauthor}}
\ead{xuzenglin@hit.edu.cn}
\address[address1]{SMILE Lab, School of Computer Science and Engineering, University of Electronic Science and Technology of China, Chengdu 610031, China.}
\address[address2]{Center for Quantum Software and Information, University of Technology Sydney, NSW 2007, Australia}
\address[address3]{Ping An Life Insurance Company of China, Ltd.}
\address[address4]{Department of Computer Science, University of Utah, Salt Lake City, 84112 Utah, USA}
\address[address5]{School of Computer Science and Engineering, Harbin Institute of Technology, Shenzhen, Shenzhen, 510085 Guangdong, China}
\address[address6]{Center for Artificial Intelligence, Peng Cheng Laboratory, Shenzhen, 510085 Guangdong, China}
\cortext[equalcontribution]{Equal Contribution}
\cortext[mycorrespondingauthor]{Corresponding Author}

	\begin{abstract}
Deep neural networks (DNNs) have achieved outstanding performance in a wide range of applications, e.g., image classification, natural language processing, etc. Despite the good performance, the huge number of parameters in DNNs brings challenges to efficient training of DNNs and also their deployment in low-end devices with limited computing resources. In this paper, we explore the correlations in the weight matrices, and  approximate the weight matrices with the low-rank block-term tensors. We name the new corresponding structure as block-term tensor layers (BT-layers), which can be easily adapted to neural network models, such as CNNs and RNNs. In particular, the inputs and the outputs in  BT-layers are reshaped into low-dimensional high-order tensors with a similar or improved representation power. Sufficient experiments have demonstrated that BT-layers in CNNs and RNNs can achieve a very large compression ratio on the number of parameters while preserving or improving the representation power of the original DNNs.

	\end{abstract}
	
\begin{keyword} 
Tensor Networks \sep Network Compression \sep Neural Networks \sep Deep Learning
\end{keyword}
\end{frontmatter}


%



\section{Introduction}\label{sec:introduction}
Deep neural networks (DNNs) have achieved significantly improved performance in a number of applications, such as image classification, image captioning, video classification, speech recognition, machine translation, etc~\cite{SCHMIDHUBER201585}. Representative DNN architectures include Convolutional Neural Networks (CNNs)~\cite{lecun1995convolutional} and Recurrent Neural Networks (RNNs)~\cite{rumelhart1988learning,sutskever2014sequence,LiuHBDBX18}, which are used in capturing the spatial and temporal information from input data, respectively.
However, due to their complex structure, the multiple layers and the huge amount of parameters, the training of DNNs becomes more difficult usually with higher space costs and more computational than classical machine learning models. For example, the winning network in the 2012 ImageNet Challenge by Krizhevsky et al. contains 60 million parameters with five
convolutional layers and three fully-connected layers~\cite{Krizhevsky2012NIPSCNN}. As a result, the training of DNNs usually costs several days even on powerful Graphics Processing Units (GPUs), and the trained model also requires a huge memory consumption~\cite{WANG2017219,Wang2018SDG}.
Meanwhile, the redundancy of parameters in DNNs brings obstacles to the training procedure, probably leading to suboptimal local optimum.
Therefore, compressing DNN architectures to reduce the parameter size is becoming an important issue in decreasing temporal and spatial complexity and reducing redundancy~\cite{cheng2017survey,lu2016learning,han2015learning}.



To obtain compact neural networks, researchers have studied efficient low-rank approximation techniques in network design, since local correlations naturally exist in natural images~\cite{lecun1995convolutional,Yu17compression}, and the dense connection is significantly inefficient at finding the spatial latent local correlation. Among the low-rank methods, tensor representations have demonstrated good performance in reducing the parameter sizes~\cite{novikov2015tensorizing,kossaifi2017tensor,Ye_2018_CVPR,Denton2014ELS}. Some research works focus on speeding up the convolutional layers of DNNs~\cite{lebedev2014speeding,Denton2014ELS}, and some
on reducing the redundency of the fully-connected layers(FC-layers) with tensor layers~\cite{novikov2015tensorizing,kossaifi2017tensor}.

In this paper, we focus on reducing the huge parameter size of the fully-connected layers in DNNs. Therefore, we resort to the representation power of Block-term Tensor decomposition (BTD) \cite{de2008decompositions2}, and reduce the fully-connected layers in DNNs as the form of Block-term Tensor layers. Since the block-term tensor decomposition is a low-rank decomposition method combining Candecom/Parafac (CP) decomposition~\cite{carroll1970analysis} and Tucker decomposition~\cite{tucker1966some}, it approximates a high-order tensor by the sum of Tucker models, which is more robust than the CP decomposition and the original Tucker decomposition. In addition, compared with the tensor train decomposition~\cite{2014AnPhyTensorNetwork}, which often suffer from the instability of ranks~\cite{zhao2016tensor}, we have demonstrated that Block-term decomposition can be more stable.




In detail, we transform  the input data $\mathbf{x}$ into a tensor of various orders, and then replace the weight matrices (i.e., $\mathbf{W}$) with new weights organized as a block-term tensor format (as will be shown in Figure \ref{fig:tensorproduct}).
While in the training phase, the sparse-connected BT layer will learn the local correlations from the input.
By integrating the new BT layer into  CNNs and RNNs architectures, we obtain new compact neural network architectures, shorted as BT-CNN and BT-RNN respectively. An illustration can be found in Fig.~\ref{fig:btcnn} and Fig.~\ref{fig:btlstm} for CNN and RNN(in particular, LSTM), respectively. Compared with their vanilla versions, the new model can have a similar representation power while with much fewer parameters. 


This paper extends the Block-term layer in LSTM~\cite{Ye_2018_CVPR} to more general neural network architectures.
The major contributions of the proposed BT layer can be concluded as follows:
\begin{itemize}
    \item The redesigned BT layer can reduce model parameters while retain the design philosophy of the original architecture. For LSTMs, the parameter reduction also speeds up the convergence, as demonstrated in Fig.~\ref{fig:loss_convergence_ucf11}.
    \item With the core tensor in the block term tensor structure, factors of the input data can have more interactions with each other, enhancing the ability to capture sufficient local correlations. Empirical results show that, compared with the standard neural network and the Tensor Train (TT) model~\cite{oseledets2011tensor}, the BT model has a better expressive ability with the same amount of model parameters, as demonstrated in Tables~\ref{e:table:acc_on_ImageNet} and \ref{tbl:ucf11_acc_state_of_the_art}.
\end{itemize}

This paper is organized as follows.  We first introduce the tensor diagrams for representing block-term tensor decomposition, followed by the corresponding BT-CNN and BT-RNN models in Section ~\ref{sec:method}. We then present evaluations of these models in Section~\ref{sec:cnn} and Section~\ref{sec:rnn}, respectively. After  reviewing and discussing recent work on tensor-based network compression methods in Section~\ref{sec:related}, we conclude our paper in Section~\ref{sec:con}.
\section{Tensorizing Matrix Product}
\label{sec:method}

The core concept of this work is to approximate the matrix vector production $\mathbf{W} \cdot \mathbf{x}$ with much fewer parameters, while still preserving the representation ability in neural networks. The technique we use for the approximation is the low-rank Block Term Decomposition (BTD), which represents $\mathbf{W} \cdot \mathbf{x}$ as a series of light-weighted small tensor products.

\subsection{Preliminaries and Background}

\subsubsection{Tensor Representation} 
\begin{figure}[t]
	\centering
	\subfigure[][Graphical representation]{\includegraphics[width=0.20\textwidth]{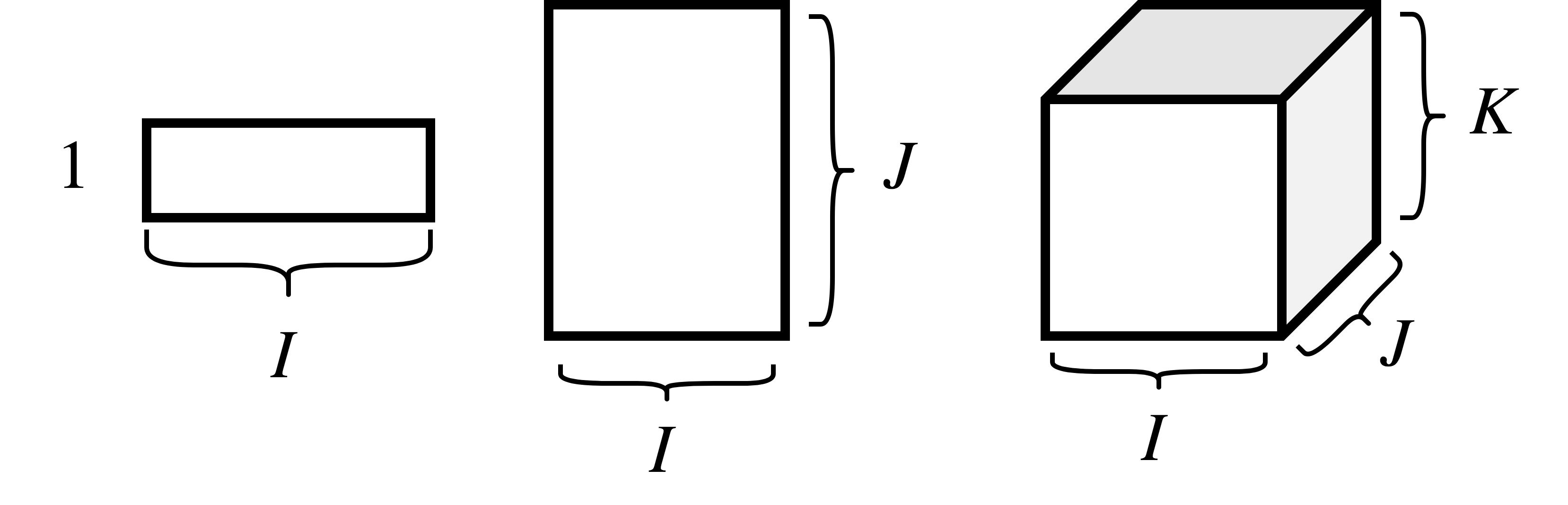} \label{fig:tensor1}} \
	\subfigure[][Symbol representation]{\includegraphics[width=0.25\textwidth]{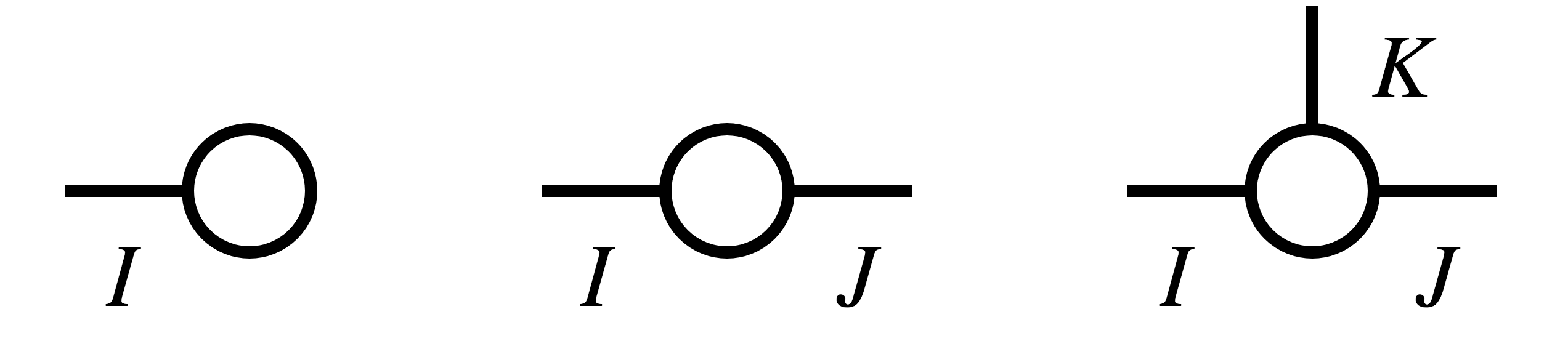} \label{fig:tensor2}}
	\caption{Two representations of vector $\mathbf{v} \in \mathbb{R}^{I}$, matrix $\mathbf{M} \in \mathbb{R}^{I\times J}$ and tensor $\bm{\mathcal{T}} \in \mathbb{R}^{I \times J \times K}$.}
	\label{fig:tensor}
\end{figure}

A tensor in neural network, also known as a multi-way array, can be viewed as a higher-order extension of a vector (i.e., an order-1 tensor) and a matrix (i.e., an order-2 tensor).  Like rows and columns in a matrix, an order-$d$ tensor $\bm{\mathcal X}\in\mathbb R^{I_1\times I_2 \ldots\times I_d}$ has $d$ modes whose lengths are represented by $I_1$ to $I_d$, respectively. The basic operations of tensors include linear arithmetic, tensor product, tensor transpose and tensor contraction. When the amount of tensors is big and the contraction relationships among their indices are complicated, a better way to represent them is using diagrams, namely tensor network diagrams. The basic symbols for tensor network diagrams are shown in Fig.~\ref{fig:tensor}, in which tensors are  denoted graphically by nodes and edges. Each edge emerged from a node denotes a mode (or order, index) \cite{cichocki2014era}. Here, we use boldface lowercase letters( e.g., $\mathbf{v}$) to denote vectors, boldface capital letters( e.g., $\bm{\mathcal W}$) to denote matrices, and boldface Euler script letters(e.g., $\bm{\mathcal A}$) to denote higher-order tensors (order-3 or higher), respectively.

\subsubsection{Tensor Contraction}
\begin{figure}[t]
	\centering
	\includegraphics[width=0.35\textwidth]{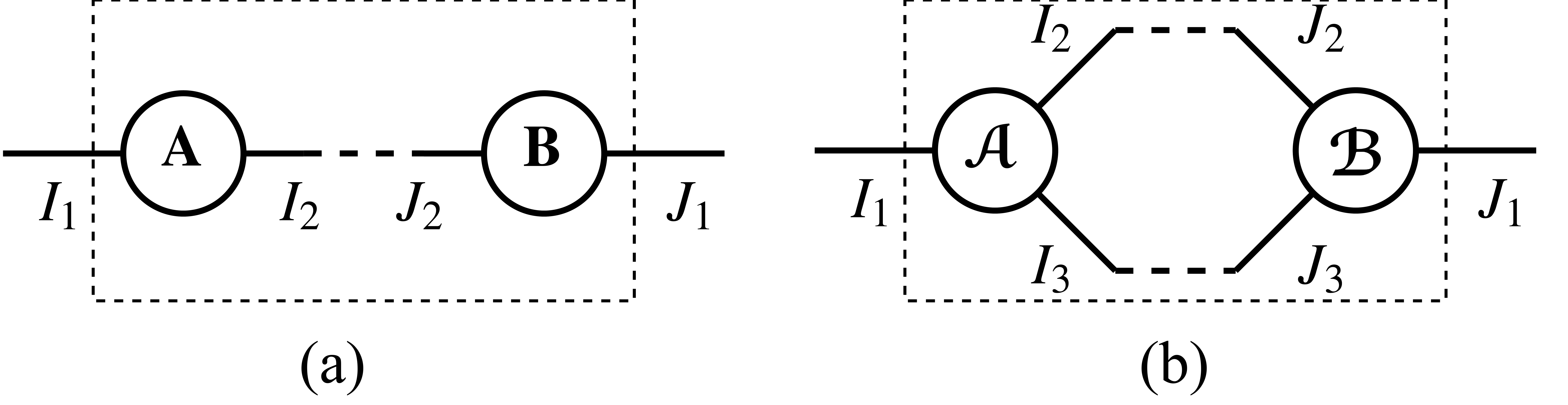}
	\caption{Diagrams of matrix product and tensor contraction.}
	\label{fig:contraction}
	\vspace{-2ex}
\end{figure}

Tensor multiplication is also known as tensor contraction.
Tensor contraction between two tensors means that they are contracted into one tensor along the associated pairs of indices as illustrated in Fig.~\ref{fig:contraction}. Since a tensor is the high dimensional extension of a matrix or vector, we can refer the tensor contraction with matrix multiplication. Giving two matrices $\bm{\mathcal A} \in \mathbb{R}^{I_1 \times I_2}$ and $\bm{\mathcal B} \in \mathbb{R}^{J_1 \times J_2}$, we can conduct the multiplication if $I_2 = J_1$ as shown in Eq.~\ref{eqt:matrix_product}.
\begin{align}
    \bm{\mathcal{C}_{i_1, j_2}} = \sum_{k=1}^{I_2} \bm{\mathcal{A}_{i_1, k}} \bm{\mathcal{B}_{k, j_2}}
    \label{eqt:matrix_product}
\end{align}

The matrix or vector multiplication is essentially the sum reduction along the same dimension. Extending this idea to high dimensional situations, we can obtain the tensor multiplication definition. For example, giving a 3-order tensor $\bm{\mathcal{A}} \in \mathbb{R}^{I_1 \times I_2 \times I_3}$ and $\bm{\mathcal{B}} \in \mathbb{R}^{J_1 \times J_2 \times J_3}$, we can conduct the tensor multiplication as follows if $I_2 = J_2$:
\begin{align}
	\bm{\mathcal{C}}_{i_1, i_3, j_1, j_3} &= \bm{\mathcal{A}} \bullet_2 \bm{\mathcal{B}} \nonumber \\
	&= \sum_{k=1}^{I_2} \bm{\mathcal{A}}_{i_1, k, i_3} \bm{\mathcal{B}}_{j_1, k, j_3}, \label{eqt:tensor_product_1}
\end{align}
where the resulted tensor $\bm{\mathcal{C}} \in \mathbb{R}^{I_1 \times I_3 \times J_1 \times J3}$ is a 4-order tensor. We use $\bullet_k$ to denote that the tensor multiplication is conducted along the $k$-th dimension.

We can also conduct multiplication along multiple dimensions. For example, if  $I_2 = J_2$ and $I_3 = J_3$, we can reduce this two dimensions as:
\begin{align}
	\bm{\mathcal{C}}_{i_1, j_1} &= \bm{\mathcal{A}} \bullet_{2,3} \bm{\mathcal{B}} \nonumber \\
	&= \sum_{k_1=1}^{I_2} \sum_{k_2=1}^{I_3} \bm{\mathcal{A}}_{i_1, k_1, k_2} \bm{\mathcal{B}}_{j_1, k_1, k_2}. \label{eqt:tensor_product_2}
\end{align}
Fig.~\ref{fig:contraction} demonstrates the matrix multiplication and tensor contraction operation in the 3-order case.

Tensor contractions among multiple tensors can be computed by performing tensor contraction between two tensors many times. Hence, the order (i.e., the total number of modes) of an entire tensor network is given by the number of dangling (free) edges that is not contracted.

\subsubsection{Block Term Decomposition (BTD)}
There are usually two basic tensor decomposition methods: the CANDECOMP/PARAFAC decomposition (CP) ~\cite{carroll1970analysis} and the Tucker decomposition~\cite{tucker1966some}. The CP method decomposes a tensor into a sum of several component rank-1 tensors; The Tucker method decomposes a tensor into a core tensor multiplied by a matrix along each mode. Their concentrations are different, as CP imposed a diagonal constraint on the core tensor of Tucker.
Thus, a more general decomposition called \textit{Block Term} (BT) decomposition, which combines CP and Tucker, has been proposed to take advantages of both of them \cite{de2008decompositions2}.

\begin{figure}[t]
	\centering
	\includegraphics[width=0.45\textwidth]{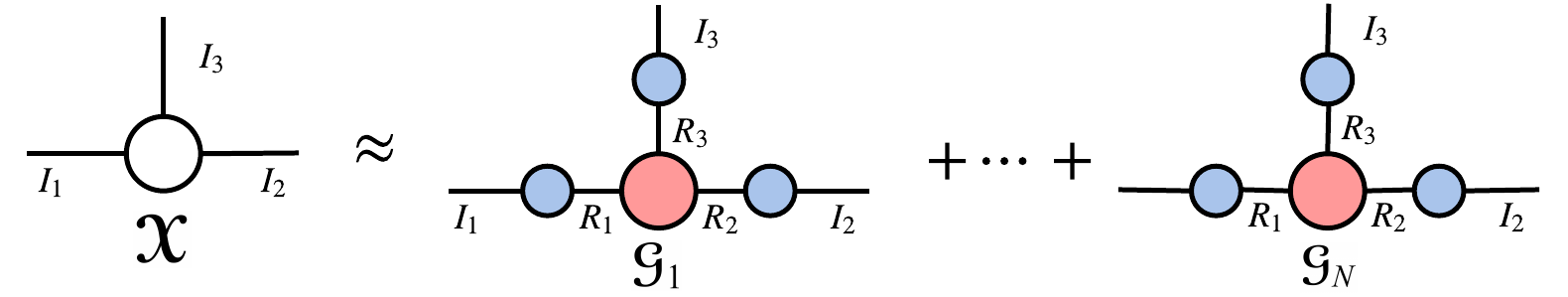}
	\caption{Illustration of the BTD in an order-3 tensor $\bm{\mathcal{X}} \in \mathbb{R}^{I_1 \times I_2 \times I_3}$.}
	\label{fig:bt-decomp}
	\vspace{-2ex}
\end{figure}

The BT decomposition aims to decompose a tensor into a sum of several Tucker models with a low Tucker-rank.
Specifically speaking, giving an order-d tensor $\bm{\mathcal X}\in\mathbb R^{I_1\times \cdots \times I_d}$, its BT decomposition can be represented by multiple Tucker models. Fig.~\ref{fig:bt-decomp} illustrates an order-3 case. In the figure,  $\bm{\mathcal G}_n \in\mathbb R^{R_1 \times \cdots \times R_d}, n \in [1, N]$ denotes the order-$d$ core tensors of the $n$-th Tucker model and each $\bm{\mathcal A}_n^{(k)} \in\mathbb R^{I_k \times R_k}, k \in [1,d]$ denotes the $k$-th factor matrix in the $n$-th Tucker model. Moreover, $\bm{\mathcal X}$ can be computed as follows:
\begin{align}
	\label{BTD_formulation}
	\bm{\mathcal{X}} = \sum_{n=1}^N \bm{\mathcal{G}}_n \bullet_1  \bm{\mathcal{A}}_n^{(1)} \bullet_2 \bm{\mathcal{A}}_n^{(2)} \bullet_3 \dots \bullet_d \bm{\mathcal{A}}_n^{(d)},
\end{align}
where $N$ denotes the CP-rank here, and $R_k$ denotes the Tucker-rank while the $d$ is the Core-order of the Tucker model.

\begin{figure}[t]
	\centering
	\subfigure[]{\label{fig:vector2tensor}
		\includegraphics[width=0.2\textwidth]{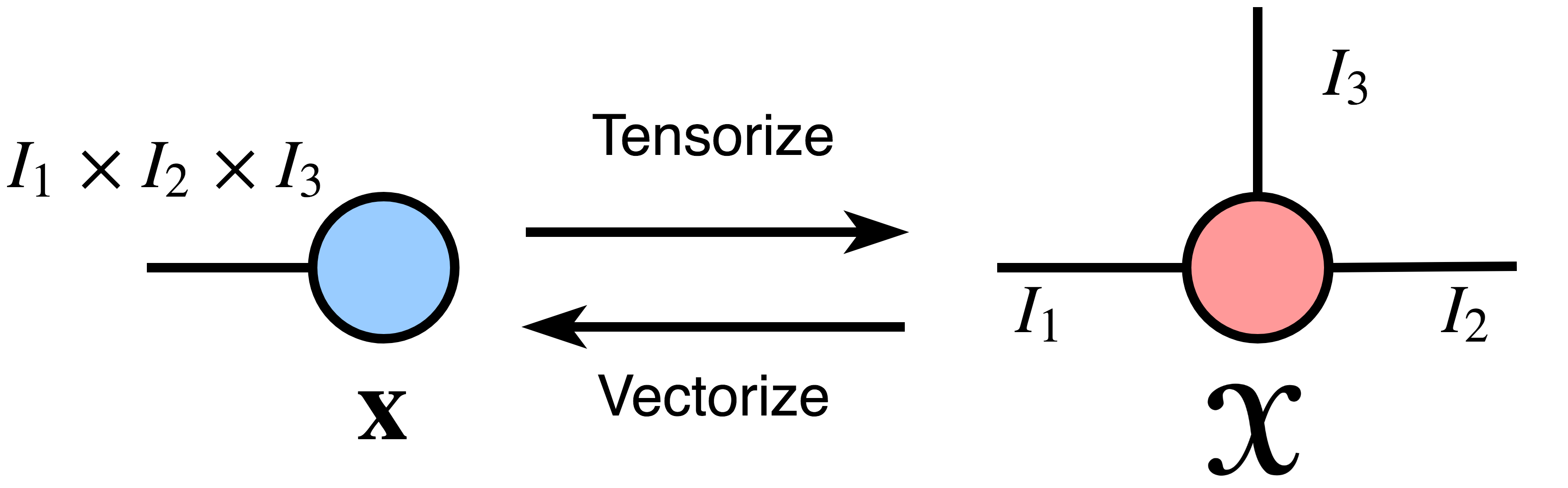}}
	\subfigure[]{\label{fig:matrix2tensor}
		\includegraphics[width=0.2\textwidth]{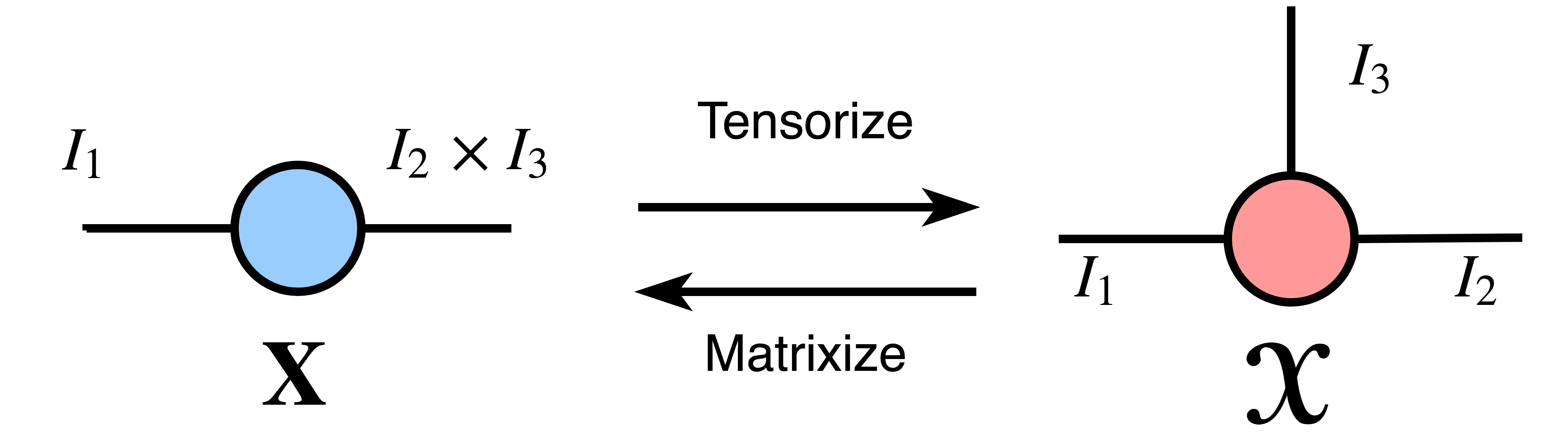}}
	\caption{Tensorization and reverse operations in a case of 3-order tensors, essentially the reshape operation.}
	\label{fig:vector_matrix_tensor}
	\vspace{-2ex}
\end{figure}

\subsection{Approximating Fully Connected Layer}
This section demonstrates how to use the BTD to approximate fully connected layer, e.g., $\mathbf{y} = \mathbf{W} \cdot \mathbf{x}$, to reduce model parameters. 1) The weight $\mathbf{W}$ and the input vector $\mathbf{x}$ should be tensorized to obtain the high order tensors $\bm{\mathcal{ W }}$ and $\bm{\mathcal{ X }}$;  2) then the weight tensor $\bm{\mathcal{ W }}$ will be reconstructed by the BTD; 3) after that, the matrix multiplication $\mathbf{W} \cdot \mathbf{x}$ can be replaced by a tensor contraction that $BTD(\bm{\mathcal{W}}) \bullet \bm{\mathcal{X}}$; 4) the reconstructed model still uses the BP algorithm~\cite{werbos1990backpropagation} to solve.

\begin{figure}[t]
	\centering
	\includegraphics[width=0.48\textwidth]{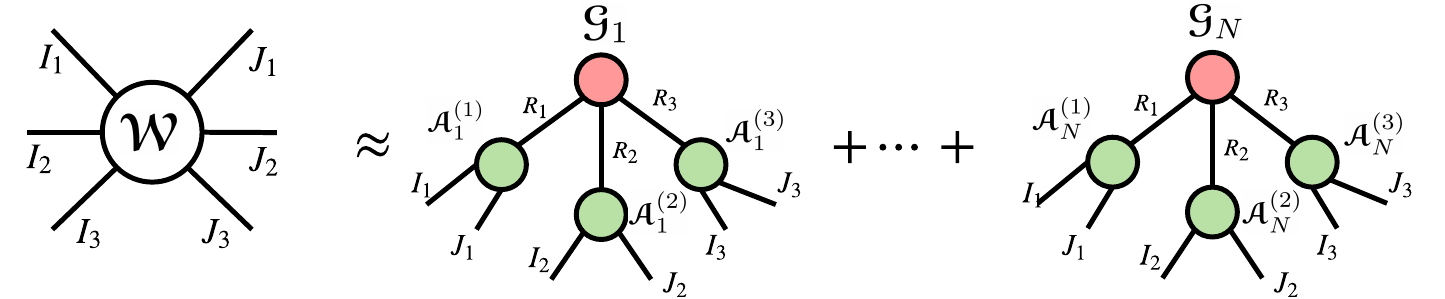}
	\caption{BTD for a 6-order weight tensor reshaped from a weight matrix $\mathbf{W} \in \mathbb{R}^{I \times J}$, where $I = I_1 \times I_2 \times I_3$ and $J = J_1 \times J_2 \times J_3$.}
	\label{fig:bt-matrix}
	\vspace{-2ex}
\end{figure}

\subsubsection{Step 1: Tensorization}
To obtain the high order tensor structure, first we should tensorize the related tensors in fully-connected layer. Tensorization is an operation that transforms a vector or matrix into a high order tensor. As illustrated in Fig.~\ref{fig:vector_matrix_tensor}, given a vector $\mathbf{x} \in \mathbb{R}^{I_1 \cdot I_2 \cdot I_3}$, we can tensorize as an order-3 tensor $\bm{\mathcal{X}} \in \mathbb{R}^{I_1 \times I_2 \times I_3}$, essentially the reshape operation, also known as tensor folding and unfolding operation.
We can also tensorize the weight matrix $\mathbf{W} \in \mathbb{R}^{J \times I}$ into a $2d$ order tensor $\bm{\mathcal{W}} \in \mathbb{R}^{J_1 \times I_1 \times J_2 \times\dots \times J_d \times I_d}$.

\subsubsection{Step 2: Applying BTD}
Next we can decompose the weight tensor by the BTD algorithm as:
\begin{align}
	BTD(\bm{\mathcal{W}}) = \sum_{n=1}^N \bm{\mathcal{G}}_n \bullet_1 \bm{\mathcal{A}}_n^{(1)} \bullet_2 \cdots \bullet_d \bm{\mathcal{A}}_n^{(d)},
	\label{eqt:bt_model}
	\vspace{-2ex}
\end{align}
here the core tensor $\bm{\mathcal{G}}_n \in \mathbb{R}^{R_1 \times \dots \times R_d}$ has $d$ dimensions and the $R_1, \ldots, R_d$ denotes the Tucker-ranks. To avoid imbalance weight sharing, in this paper we set $R = R_1 = R_2 = \cdots = R_d$. The factor tensor $\bm{\mathcal{A}}_n^{(d)} \in \mathbb{R}^{I_d \times J_d \times R_d}$ is corresponding to the input and core tensor.
To get the low-rank decomposition, we have the subjection of $R_k \leqslant I_k$ (and $J_k$), $k=1,\ldots, d$.

\subsubsection{Step 3: Substitution}
\begin{figure}[t]
	\centering
	\includegraphics[width=0.40\textwidth]{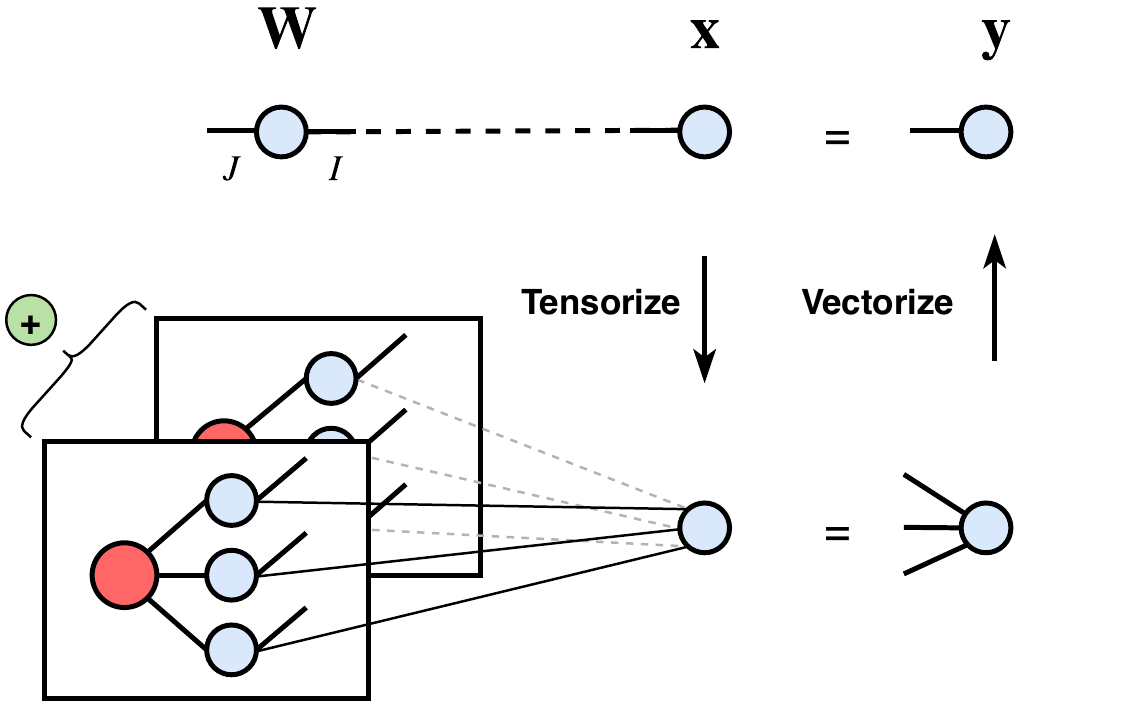}
	\caption{Diagrams of BT representation for matrix-vector product $\mathbf{y} = \mathbf{Wx}, \mathbf{W} \in \mathbb{R}^{J \times I}$. We substitute the weight matrix $\mathbf{W}$ by the BT representation, then tensorize the input vector $\mathbf{x}$ to a tensor with shape $I_1 \times I_2 \times I_3$. After operating the tensor contraction between BT representation and input tensor, we get the result tensor in shape $J_1 \times J_2 \times J_3$. With the vectorize operation, we get the output vector $\mathbf{y} \in \mathbb{R}^{J_1 \cdot J_2 \cdot J_3}$. }
	\label{fig:tensorproduct}
	\vspace{-2ex}
\end{figure}

After tensorizing the input vector and weight matrix, and decomposing the weight tensor, now we can conduct a tensor contraction operation between input tensor and BTD model as follows:
\begin{align}
	\phi(\mathbf{W}, \mathbf{x}) = BTD(\bm{\mathcal{W}}) \bullet_{1,2,\ldots,d} \bm{\mathcal{X}},
	\label{eqt:bt_representation}
\end{align}
$\bullet_{1,2,\ldots,d}$ denotes that the contraction will be conducted along all $d$ dimensions. To distinguish the new operation from the fully-connected layer, we call the proposed layer as the Block Term Layer (\textit{BTL}). Fig.~\ref{fig:tensorproduct} demonstrates the substitution intuitively.

\subsubsection{Step 4: Training BTL}
In deep neural networks, the most common training method is the Back Propagation (BP) algorithm which back propagates the error in an reverse order with the forward computation and updates the learning parameters iteratively.

The BTL training still follows the standard error back-propagation algorithm, here we derive the weight gradient and input gradient as follows:
\begin{align}
	\frac{\partial L}{\partial \bm{\mathcal{A}}_n^{(k)}} = &
	\frac{\partial L}{\partial \bm{\mathcal{Y}}}
	\bullet_k \bm{\mathcal{X}}_t \bullet_1 \bm{\mathcal{A}}_n^{(1)} \bullet_2 \dots \label{eqt:partial_BTL_A}                                                                                                                                                     \\
	                                                       & \bullet_{k-1} \bm{\mathcal{A}}_n^{(k-1)} \bullet_{k+1} \bm{\mathcal{A}}_n^{(k+1)} \bullet_{k+2} \dots \nonumber                                                                                  \\
	                                                       & \bullet_d \bm{\mathcal{A}}_n^{(d)} \bullet_{1,2,\dots,d} \bm{\mathcal{G}}_n, \nonumber                                                                                                           \\
	\frac{\partial L}{\partial \bm{\mathcal{G}}_n} =       & \bm{\mathcal{X}}_t \bullet_1 \bm{\mathcal{A}}_n^{(1)} \bullet_2 \dots \bullet_d \bm{\mathcal{A}}_n^{(d)} \bullet_{1,2,\dots,d} \frac{\partial L}{\partial \bm{\mathcal{Y}}} \label{eqt:partial_BTL_G}.
\end{align}

\section{\textit{BTL} Implementations in CNNs and RNNs}

\subsection{BT-CNN}
In classical Convolutional Neural Networks, input features from stacked convolution layers are fed into  fully-connected layers and activation layers, which serve as a non-linear classifier. To verify the efficient performance of the block-term layer, we replace the fully-connected layers with block-term layers in CNNs to construct a low-rank classifier.

\begin{figure}[t]
	\centering
	\includegraphics[width=0.45\textwidth]{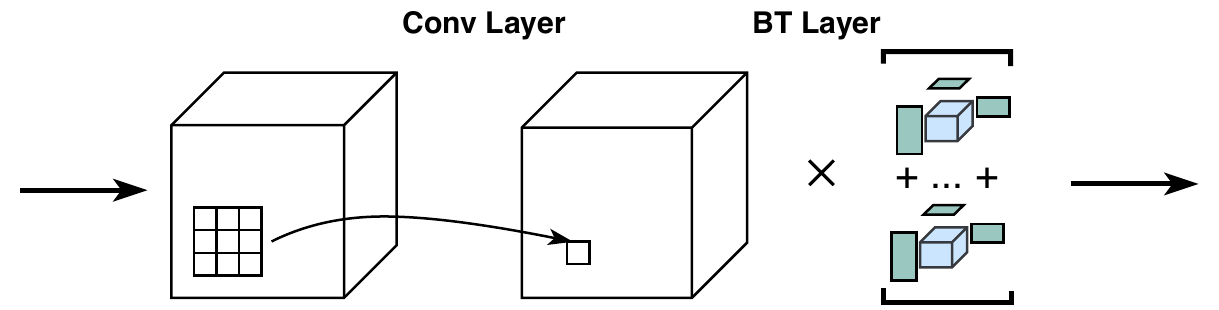}
	\caption{Architecture of BT-CNN. The redundant dense connections between input and hidden state is replaced by low-rank BT representation.}
	\label{fig:btcnn}
\end{figure}

As illustrated in Fig.~\ref{fig:btcnn}, we only replace the redundant fully connected layers while keeping the other layers. We will discuss the compression ratio in the experiment section.

\subsection{BT-LSTM}

\begin{figure}[t]
	\centering
	\includegraphics[width=0.45\textwidth]{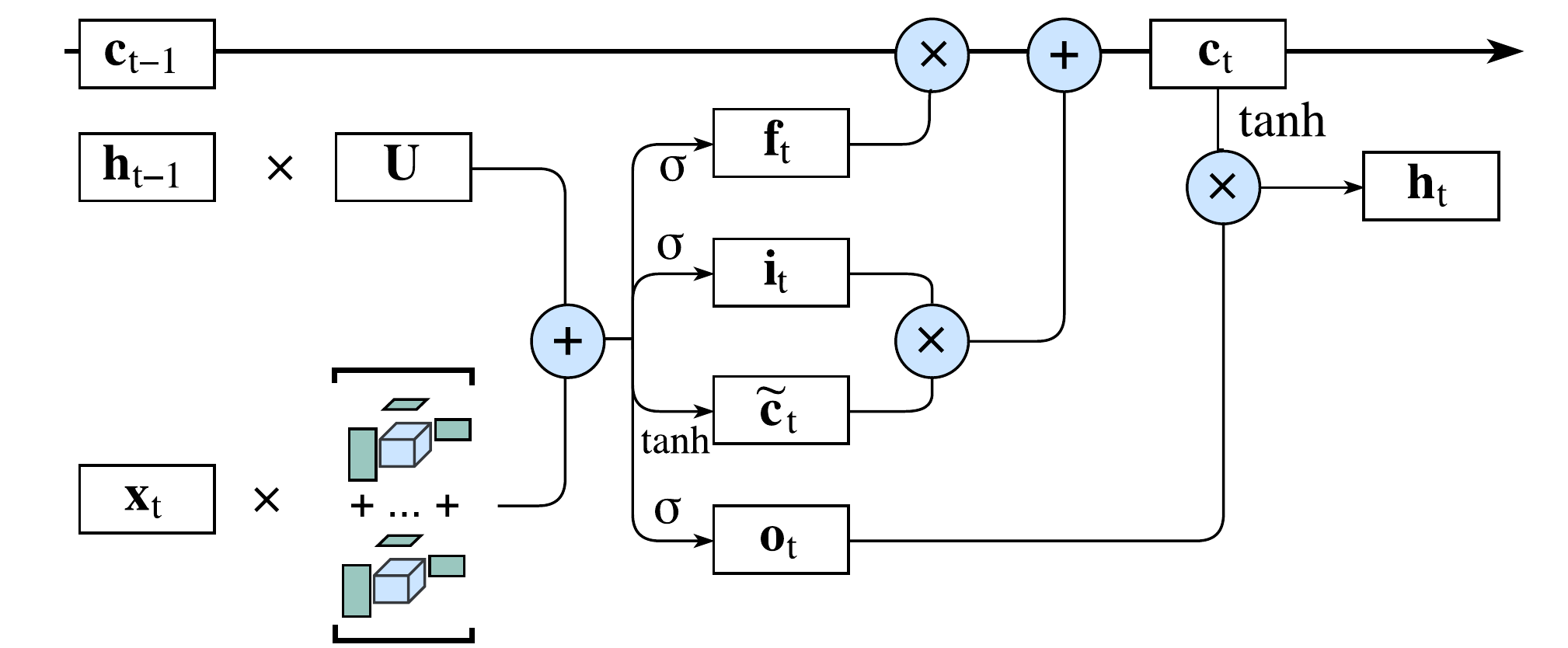}
	\caption{Architecture of BT-LSTM. The redundant dense connections between input and hidden state is replaced by low-rank BT representation.}
	\label{fig:btlstm}
\end{figure}

We demonstrate the BT-Layer in LSTM since it is the most common used RNNs . Although our discussion mainly focuses on LSTM, the proposed method can easily be generalized into other RNN variants such as Gated Recurrent Unit (GRU) by following above procedures.

When dealing with high dimensional input data, the fully connected layer between input and hidden state is extremely low-rank and taking the most training parameters in the model. 
Since operations in $\mathbf{f}_t$, $\mathbf{i}_t$, $\mathbf{o}_t$ and $\tilde{\mathbf{c}}{_t}$ have the similar computation procedure, we merge them together by concatenating $\mathbf{W}_f$, $\mathbf{W}_i$, $\mathbf{W}_o$ and $\mathbf{W}_c$ into one giant $\mathbf{W}$, and so does $\mathbf{U}$. As illustrated in Fig.~\ref{fig:btlstm}, we can simplify the LSTM formulations as follows:

\begin{align}
	\left({\mathbf{f}_t}', {\mathbf{i}_t}', {\tilde{\mathbf{c}}_t}', {\mathbf{o}_t}'\right) & = \phi(\mathbf{W}, \mathbf{x}_t) + \mathbf{U} \cdot \mathbf{h}_{t-1} + \mathbf{b} \label{eqt:concat_lstm},                 \\
	\left(\mathbf{f}_t, \mathbf{i}_t, \tilde{\mathbf{c}}_t, \mathbf{o}_t)\right)            & = \left(\sigma({\mathbf{f}_t}'), \sigma({\mathbf{i}_t}'), \tanh({\tilde{\mathbf{c}}_t}'), \sigma({\mathbf{o}_t}')\right), \\
	\mathbf{c}_t                                                                            & = \mathbf{f}_t \odot \mathbf{c}_{t-1} + \mathbf{i}_t \odot \tilde{\mathbf{c}_t},                                           \\
	\mathbf{h}_t                                                                            & = \mathbf{o}_t \odot \tanh(\mathbf{c}_t),
\end{align}
where the $\phi(\mathbf{W}, \mathbf{x}_t)$ denotes the low rank BT layer.

\subsection{\# Params w.r.t. Hyper-parameters}
BTD decomposes a high order tensor into several low-rank Tucker blocks to reduce the parameters. Here we give out the total number of parameters of BTD:
\begin{align}
	P_{BTD} = N (\sum_{k=1}^d I_k J_k R + R^d).
	\vspace{-2ex}
	\label{eqt:number_of_parameters}
\end{align}
The model parameters of the fully-connected layer mainly fall into the weight matrix $\mathbf{W}$, whose the \#Params is $P_{FC} = I\times J = \prod_{k=1}^d I_k J_k$. To simplify the comparison we define the compression ratio, which is essentially the ratio of \#Params between the BT model and FC model:
\begin{align}
	\text{Comp.R} = \frac{P_{FC}}{P_{BTD}}
\end{align}

\begin{figure}[t]
	\centering
	\includegraphics[width=0.45\textwidth]{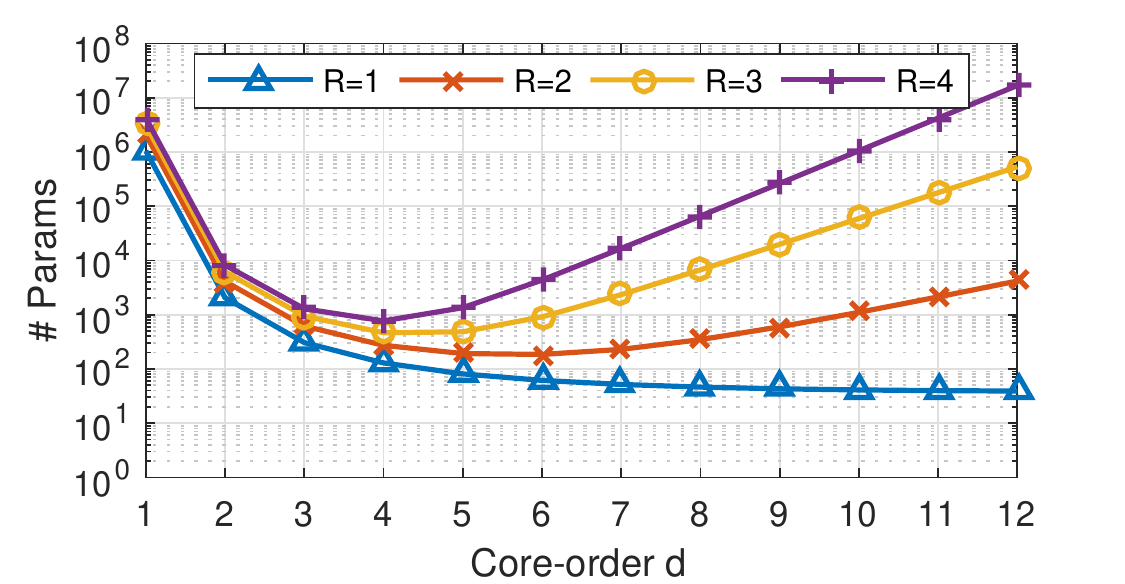}
	\caption{The number of parameters w.r.t Core-order $d$ and Tucker-rank $R$, in the setting of $I = 4096, J = 256, N = 1$. While the vanilla RNN contains $I\times J = 1048576$ parameters. Refer to Eq. (\ref{eqt:number_of_parameters}), when $d$ is small, the first part $\sum_1^d I_k J_k R$ does the main contribution to parameters. While $d$ is large, the second part $R^d$ does. So we can see the number of parameters will go down sharply at first, but rise up gradually as $d$ grows up (except for the case of $R=1$).}
	\label{fig:paramd}
	\vspace{-2ex}
\end{figure}

To demonstrate how total amount of parameters vary w.r.t different hyper-parameter settings, we plot the \#Params w.r.t. the Core-order $d$ and Tucker-rank $R$ in Fig.~\ref{fig:paramd}. 

The Core-order denotes the order of the input tensor and core tensors. When $d=1$ the BT model will fallback to the FC model. As illustrated in Fig.~\ref{fig:paramd}, a proper choice of the core-order is relative important. In our practice, we choose $d \in [2,5]$.

The concept of Tucker-rank $R$ is similar to the rank of Singular Value Decomposition (SVD), which is the dimension of the core tensor. From Eq.(\ref{eqt:number_of_parameters}), the \#Params will increase exponentially with the increase of $R$. To obtain a low-rank model, we have $R \leqslant I_k, k \in [1,d]$. While the $I_k \approx \sqrt{d}{I}$, we have the limited choice of $R$, which alleviates the difficulty of the hyper-parameter setting.

The introduction the the CP-rank $N$ is the mainly different between Tucker decomposition and BT decomposition, which effectively widens the network in a linear manner. 

\subsection{Complexity Analysis}
Eq. (\ref{eqt:bt_model}) follows the same computation order with the FC layer, resulting in high computation cost as $\mathcal{O}(IJR)$ which falls in the last tensor contraction. 

Eq. (\ref{eqt:bt_model}) raises the computation peak, $\mathcal{O}(IJR)$, at the last tensor product $\bullet_d$, according to left-to-right computational order. However, we can reorder the computations to further reduce the total model complexity $\mathcal{O}(NdIJR)$. The reordering is:
\begin{align}
	\phi(\mathbf{W}, \mathbf{x}_t) = \sum_{n=1}^N \bm{\mathcal{X}}_t \bullet_1 \bm{\mathcal{A}}_n^{(1)} \bullet_2 \dots \bullet_d \bm{\mathcal{A}}_n^{(d)} \bullet_{1,2,\ldots,d} \bm{\mathcal{G}}_n. \label{eqt:bt_reorder}
\end{align}
In Eq.(\ref{eqt:bt_reorder}), the computation raised at the last contraction is $\bm{\mathcal{O}}(IJ_{max} R^d)$. Since we have $J_{max} R^d \leqslant J$ and $R \leqslant J_{min}$, the complexity in the new computation order will be always smaller than the previous. Finally we reduce the forward computation complexity from $\mathcal{O}(Nd I J R)$ to $\mathcal{O}(Nd I J_{max} R^d)$, which is more endurable.

We derive the gradients in backward phase and ignores the dimension alignment for simplicity:
\begin{align}
	\frac{\partial L}{\partial \bm{\mathcal{A}}_n^{(k)}} = &
	\frac{\partial L}{\partial \bm{\mathcal{Y}}}
	\bullet_k \bm{\mathcal{X}}_t \bullet_1 \bm{\mathcal{A}}_n^{(1)} \bullet_2 \dots \label{eqt:partial_A}                                                                                                                                                     \\
	                                                       & \bullet_{k-1} \bm{\mathcal{A}}_n^{(k-1)} \bullet_{k+1} \bm{\mathcal{A}}_n^{(k+1)} \bullet_{k+2} \dots \nonumber                                                                                  \\
	                                                       & \bullet_d \bm{\mathcal{A}}_n^{(d)} \bullet_{1,2,\dots,d} \bm{\mathcal{G}}_n, \nonumber                                                                                                           \\
	\frac{\partial L}{\partial \bm{\mathcal{G}}_n} =       & \bm{\mathcal{X}}_t \bullet_1 \bm{\mathcal{A}}_n^{(1)} \bullet_2 \dots \bullet_d \bm{\mathcal{A}}_n^{(d)} \bullet_{1,2,\dots,d} \frac{\partial L}{\partial \bm{\mathcal{Y}}} \label{eqt:partial_G}.
\end{align}
Both Eq.(\ref{eqt:partial_A}) and Eq.(\ref{eqt:partial_G}) follow the similar computation pattern with Eq.(\ref{eqt:bt_representation}), raising the same complexity in $\mathcal{O}(d I J_{max} R^d)$. 

\begin{table}[t]
	\caption{The complexity in forward and backward computation of the fully-connected layer, TT layer and BT layer.}
	\begin{center}
		\begin{tabular}{l|l|l}
			\hline
			Method          & Time                               & Memory               \\
			\hline\hline
			FC forward     & $\mathcal{O}(IJ)$                  & $\mathcal{O}(IJ)$    \\
			FC backward    & $\mathcal{O}(IJ)$                  & $\mathcal{O}(IJ)$    \\
			TT forward  & $\mathcal{O}(d I R^2 J_{max})$     & $\mathcal{O}(R I)$   \\
			TT backward & $\mathcal{O}(d^2 I R^4 J_{max})$   & $\mathcal{O}(R^3 I)$ \\
			BT forward  & $\mathcal{O}(N d I R^d J_{max})$   & $\mathcal{O}(R^d I)$ \\
			BT backward & $\mathcal{O}(N d^2 I R^d J_{max})$ & $\mathcal{O}(R^d I)$ \\
			\hline
		\end{tabular}
	\end{center}
	\label{tbl:complexity}
	\vspace{-2ex}
\end{table}

The detail of the computation complexity in various architectures is shown in Table \ref{tbl:complexity}.

\section{Evaluation on BT-CNN}
\label{sec:cnn}

The goal of this experiment is to verify the effectiveness of our proposed BT-nets using various neural network architectures in terms of the compression ratio and accuracy of BT-layers.
For fair comparison, we adopt pre-fixed architectures of CNNs available in various platforms (e.g., Tensorflow and MxNet) as our test beds, and  replace only the  FC-layers with BT-layers.
As tensor train methods   have achieved better performance than SVD in compressing CNNs (as shown in ~\cite{novikov2015tensorizing}), we focus to compare our proposed method with tensor train methods.
For consistency, we adopt three three widely-used benchmark datasets,i.e., MNIST \cite{lecun1998mnist}, Cifar10 \cite{krizhevsky2009learning}, and ImageNet \cite{deng2012imagenet}.



The experiments are performed on a server with one NVIDIA TITAN Xp GPU.
For fair comparison, we train all networks from scratch with stochastic gradient descent with momentum of 0.9. To avoid the influence of random initialization and the problem of gradient vanishing or exploding, it is necessary to add a batch normalization layer after the BT-layer.

\subsection{Results on MNIST}
As a baseline we use the convolutional neural network LeNet-5
\cite{lecun1998gradient}, with two convolutional (plus activation function and max-pooling) layers followed by two fully-connected layers (FC-layers) of size $800\times 500$ and $500\times 10$.

We fix the convolutional part of the original network and only replace the first FC-layer with the BT-layer.
The BT-layer reshapes the input and output tensors as $5\times 5\times 8\times 4$ and $5\times 5\times 5\times 4$ tensors respectively. As the prediction task for MNIST  is quite easy, we simply use one block term decomposition (which is  Tucker decomposition and $N=1$) and vary the Tucker-rank from 2 to 3 ($R=2$ or $R=3$). Here we use $BT-N:1-R:2$ to denote a BT network with the CP rank equal to 1 and the Tucker-rank equal to 2, which is also shorted as $BT-R:2$ when $N=1$. We run the TT-net~\cite{novikov2015tensorizing} as a competitor by replacing the first FC-layer with the TT-layer. For a fair comparison with a similar amount of parameters, we set the TT-rank to 2, and the corresponding network is denoted by $TT-R:2$.

Table~\ref{e:table:acc_on_mnist} reports the results on the MNIST dataset. 
It can be observed that the parameter amount is greatly reduced while with a slight improvement in accuracy.
The first column represents the different network architectures, the middle two columns represent the number of parameters in the first FC-layer (or its alternatives) and the compression ratio respectively. The last column represents the accuracy on the test set. We can see at first glance that the number of parameters in the FC-layer can be reduced from 800$\times$500 to 228 in ``BT-R:2'' and the compression ratio can up to 1754, with almost the same accuracy. The compression ratio of the entire network is about 14.
We can also observe that ``BT-R:3'', with 399 parameters in BT-layer, has the same accuracy as the baseline while TT-net lost 0.03\% in performance on about the same order of magnitude of the parameter amount.


\begin{table}[t]
	\centering
	\caption{Results on MNIST with various Tucker Rank settings. By substituting the first fully connected layer, we got great performance in both TT and BT networks.}
	\begin{tabular}{c|ccc}
		\hline
		Architecture & \# params      & Comp.R        & Acc (\%)       \\
		\hline \hline
		Baseline     & 800$\times$500 & 1             & 99.17          \\
		TT-R:2       & 342            & 1169          & 99.14          \\
		BT-R:2       & \textbf{228}   & \textbf{1754} & 99.14          \\
		BT-R:3       & 399            & 1002          & \textbf{99.18} \\
		\hline
	\end{tabular}
	\label{e:table:acc_on_mnist}
\end{table}


\subsection{Results on CIFAR10}


We refer to the tensorflow implantation\footnote{https://github.com/tensorflow/models/tree/master/tutorials\\/image/cifar10} as the baseline CNN implementation, which consists of two Convolutional, Local Respond Normarlization (LRN) and Max-pooling layers followed by three FC-layers of size $2304\times 384$, $384\times 192$ and $192\times 10$, respectively.

We similarly replace the first FC-layer with BT-layer which reshapes the input and output dimensions as $6\times 6\times 8\times 8$ and $6\times 4\times 4\times 4$ respectively. TT-net  replaces the first FC-layer with TT-layer which has the same output dimension as BT-layer.
For fair comparison, we let the CP-rank vary from 1 to 8 and the Tucker-rank vary from 1 to 3 in the BT-layer, and let the TT-rank equal 2 and 8 in the TT-layer.

Some results of the CIFAR10 dataset are reported in Table~\ref{e:table:acc_on_cifar10}, and others can be found in Fig.~\ref{e:fig:BTrank}. We can see that when using the ``BT-N:1-R:2'' structure, the compression ratio is up to 3351 at the cost of about 1\% reduction in accuracy. By comparison, the compression ratio of ``TT-R:2'' is only 2457 with almost the same accuracy as the BT-layer. In response to the increase in the architecture's complexity, we observe that ``BT-N:4-R:3'' has a larger compression ratio while obtaining a better accuracy at the same time compared with ``TT-R:8''.  ``BT-N:4-R:3'' can have a total compression  ratio of 2.98.

\begin{table}[t]
	\centering
	\caption{Results on CIFAR10 with various architectures.}
	\begin{tabular}{c|ccc}
		\hline
		Architecture & \# params       & Comp.R        & Acc (\%)       \\
		\hline \hline
		Baseline     & 2304$\times$384 & 1             & \textbf{85.99} \\
		TT-R:2       & 360             & 2457          & 84.90          \\
		TT-R:8       & 4128            & 214           & 85.70          \\
		BT-N:1-R:2   & \textbf{264}    & \textbf{3351} & 84.95          \\
		BT-N:4-R:2   & 1056            & 838           & 85.47          \\
		BT-N:4-R:3   & 1812            & 488           & 85.83          \\
		\hline
	\end{tabular}\label{e:table:acc_on_cifar10}
\end{table}

\subsection{Results on ImageNet}
The ILSVRC 2012 (ImageNet) is a large dataset which consists of 1.2 million images for training and 50,000 for validation and comprises 1000 classes. As a baseline, we use the AlexNet architecture\footnote{https://github.com/apache/incubator-mxnet/tree/master\\/example/image-classification}, which has three FC-layers of size $6400\times 4096$, $4096\times 4096$ and $4096\times 1000$.

We replace the first FC-layer with BT-layer where the input and output dimensions are reshaped as $10\times 10\times 8\times 8$ and $8\times 8\times 8\times 8$ respectively. The same dimension  reshaping is performed in TT-layer as well. As in the CIFAR10 case, we experiment with two groups of variations (simple and complex) of the BT-layer and the TT-layer. Accordingly, we choose ``BT-N:1-R:2'' and ``BT-N:4-R:2'' as BT-layers and set TT-rank as 2 and 8 in TT-layers.

In Table~\ref{e:table:acc_on_ImageNet} we report the compression ratio, Top-1 and Top-5 accuracy on different architectures. From the results we see that BT-layer in the best case (``BT-N:4-R:2'') can get a  compression ratio of 11070 (from 6400$\times$4096 parameters to 2368) on amount of parameters while achieving a slightly better Top-1 and Top-5 accuracy than baseline at the same time. The total compression ratio of the entire network is 2.06. By comparison, ``TT-R:8'' only gets a  compression ratio of 2528 and even about 1\% accuracy drop. Similarly, ``BT-N:1-R:2'' gets a compression factor of more than 40,000 with 2.2\% decrease in Top-5 accuracy, better than ``TT2''.
Please note that all the experiments on the Imagenet are performed without fine-tuning.


\begin{table}[t]\caption{Results on ImageNet with various architectures.}
	\centering
	\begin{tabular}{c|ccc}
		\hline
		Architecture & Comp.R         & Top-1 Acc (\%) & Top-5 Acc (\%) \\
		\hline \hline
		Baseline     & 1              & 56.17          & 79.62          \\
		TT-R:2       & 30340          & 52.14          & 76.40          \\
		TT-R:8       & 2528           & 55.11          & 78.61          \\
		BT-N:1-R:2   & \textbf{44281} & 53.20          & 77.38          \\
		BT-N:4-R:2   & 11070          & \textbf{56.48} & \textbf{79.69} \\
		\hline
	\end{tabular}\label{e:table:acc_on_ImageNet}
\end{table}





\begin{table}[t]
\caption{Results of adding FC, TT and BT layers to ResNet-20 on cifar10}
\label{e:table:acc_on_resnet}
\vspace{-2ex}
\begin{center}
\begin{small}
\begin{tabular}{l|cc}
\hline
 Architecture & \# Parameters & Acc (\%)   \\
\hline\hline
  Baseline & -    & 91.73 \\
  add FC & 1024$\times$1024    & 90.69 \\
  add TT2 & 480   & 88.97 \\
  add TT8 &  5760  &  90.30 \\
  add BT-N:1-R:2 & \textbf{336}  & 90.58 \\
  add BT-N:4-R:2 & 1344   & \textbf{92.13} \\
\hline
\end{tabular}
\end{small}
\end{center}
\end{table}

\subsection{Extended Experiment}

\subsubsection{Extension to Convolutional Layers}
The weight in a convolutional layer (conv-layer) is usually denoted by an order-4 tensor $\bm{\mathcal{W}} \in \mathbb{R}^{H \times W \times C_{in} \times C_{out}}$, where $H$ and $W$ denotes the height and width of the perception area, and $C_{in}$ and $C_{out}$ denotes the size of input channels and output channels.
Since the first three indices are representing the input and the last one represents the output, we can view the weight tensor $\bm{\mathcal{W}}$ as a matrix $\bm{W}\in \mathbb{R}^{HW C_{in}\times C_{out}}$, then we perform the generalized BT decomposition on it as stated above. Hence, we could transform conv-layer to BT-layer as well.

However, due to the conv-layer's specific characteristics, we should pay attention to the following:
\begin{itemize}
	\item As the matrix $\bm{W}\in \mathbb{R}^{HWC_{in}\times C_{out}}$ is inherently belonging to a tensor, we require dividing the dimensions according to the original dimensions, for example $H\times W\times C_{in}^1\times C_{in}^2$ and $C_{out}^1\times C_{out}^2$, to maintain the high-order structure information of the convolutional layers.
	\item In general, $H$ and $W$ are far less than $C_{in}$ and $C_{out}$, so the key is dividing the larger $C_{in}$ and $C_{out}$ suitably. In practice, we divide the input and output dimensions as $H\times W\times C_{in}^1\times C_{in}^2\cdots$ and $1\times 1\times C_{out}^1\times C_{out}^2\cdots$, respectively.
\end{itemize}

\subsubsection{Results of Extensions to Conv-layer}
Since the CIFAR-Net\footnote{https://github.com/tensorflow/models/blob/master/research/\\slim/nets/cifarnet.py} has the largest parameter occupation rate of conv-layers among the above three networks, we evaluate the results of extensions to conv-layer on the CIFAR10 dataset. In CIFAR-Net, the weight in the second conv-layer, which contains the most parameters, which is $5\times 5\times 64\times 64$. We replace it with a BT-layer where the dimensions are reshaped as $5\times 5\times 64$ and $1\times 1\times 64$,  respectively. We set the CP-rank as 2, Tucker-rank as 3, and the corresponding TT-rank as 6. For a better comparison, we optionally substitute the two layers. The results are illustrated in Table~\ref{e:table:acc_conv_fc}. We can see that ``BT-N:2-R:3 + FC'', which only replaces the conv-layer with a BT-layer, obtains the best accuracy since BT gives a better representation of the weight of the conv-layer. In addition, ``BT-N:2-R:3 + BT-N:2-R:3'' could get a total Comp.R of 10.10 with only about 0.5\% reduction in accuracy, while ``TT-R:6 + TT-R:6'' reduces about 1\% in accuracy with a total Comp.R of 9.87.

\begin{table}[h]
	\caption{Results of compressing conv-layer and FC-layer on CIFAR10 dataset}
	\label{e:table:acc_conv_fc}
	\vskip 0.1in
	\begin{center}
		\begin{small}
				\begin{tabular}{c|cc}
					\hline
					Architecture            & Total Comp.R & Acc (\%)       \\
					\hline\hline
					Conv + FC               & 1.00         & 86.35          \\
					TT-R:6 + FC             & 1.08         & 86.50          \\
					Conv + TT-R:6           & 5.76         & 85.99          \\
					TT-R:6 + TT-R:6         & 9.87         & 85.36          \\
					BT-N:2-R:3 + FC         & 1.09         & \textbf{86.54} \\
					Conv + BT-N:2-R:3       & 5.83         & 86.10          \\
					BT-N:2-R:3 + BT-N:2-R:3 & 10.10        & 85.79          \\
					\hline
				\end{tabular}
		\end{small}
	\end{center}
	\vskip -0.1in
\end{table}

\subsubsection{Results of applying BT-layers to the U-Net structures}
Since the U-Net is a relatively small-scaled convolutional neural network which includes multiple convolutional and deconvolutional layers, we evaluate the  results of applying BT-layers to the U-Net.  As a baseline, we use the U-net structure and conduct the detection task of radio frequency interference (RFI) in radio astronomy as the same as \cite{akeret2017radio}.  We replace the last conv-layer in the U-net with the BT-layer, because the last conv-layer contains the most parameters, which is $3\times3\times256\times256$. The results are illustrated in Table~\ref{e:table:acc_Unet}. We can see that applying the BT-layer(``BT-N:1-R:2'') could get could get a compression ratio of 4.50 with only about 0.8\% reduction in accuracy.
BT-layer in the another case (“BT-N:2-R:3”) can get a compression ratio of 1.50 on amount of parameters while achieving slightly better accuracy than baseline.

\begin{table}[h]
\caption{Results of applying TT-layers and BT-layers to U-Net}
\label{e:table:acc_Unet}
\vspace{-2ex}
\begin{center}
\begin{small}
\begin{tabular}{l|cc}
\hline
 Architecture & Compression Ratio & Acc (\%)   \\
\hline\hline
  Baseline & 1   & 84.2 \\
  TT2 & 4.50  & 79.9 \\
  TT4 & 2.49   &  83.7 \\
  BT-N:1-R:2 & \textbf{4.50}  & 83.4 \\
  BT-N:2-R:3 & 1.50   & \textbf{84.9} \\
\hline
\end{tabular}
\end{small}
\end{center}
\end{table}

\subsubsection{Comparison between pyramid-structure CNN and BT-layers}
As described in~\cite{DBLP:journals/corr/UllahP16}, the pyramid-structure CNN starts from a big first layer and then refines the features at each higher layer, until it achieves a reduced most discriminative set of features like a pyramid. We change the filter numbers of CIFAR-Net to construct a pyramid-structure CNN and replace the first FC-layer with the BT-layer. The results are illustrated in Table~\ref{e:table:acc_Pyramid-structure}. As the same as ~\cite{DBLP:journals/corr/UllahP16}, here we use "C10: 64-64-384-192-10" to denote a CNN which has 64 filters in the first convolutional layer and 64 filters in the second convolutional layer followed by several fully-connected layers. 'SPryC10' denotes a pyramid-structure CNN. Comparing "C10 + BT-N:2-R:3" with "SPyrC10", we can see that our method achieves better results with much fewer parameters. In addition, BT-layer can even compress the parameters further with only about 1\% reduction in accuracy, when we compare "SPyrC10 + BT-N:2-R:3" and "SPyrC10". These experiments demonstrate the wide applicability of the BT-layers.

\begin{table}[h]
\caption{Results of comparing between pyramid-structure CNN and BT-layers}
\label{e:table:acc_Pyramid-structure}
\vspace{-2ex}
\begin{center}
\begin{small}
\begin{tabular}{l|cc}
\hline
 Architecture & Total Parameters & Acc (\%)   \\
\hline\hline
  C10 = 64-64-384-192-10 & 1,067,712   & \textbf{86.35} \\
  SPyrC10 = 64-36-384-192-10  & 635,812  & 85.60 \\
  C10 + BT-N:2-R:3 & 183,882  & 86.10 \\
  SPyrC10 + BT-N:2-R:3 & \textbf{138,958}   & 84.67 \\
\hline
\end{tabular}
\end{small}
\end{center}
\end{table}

\subsubsection{Sensitive Analysis of BT-ranks and the Tensor-order}

\begin{figure}[t]
	\centering
	\includegraphics[width=0.4\textwidth]{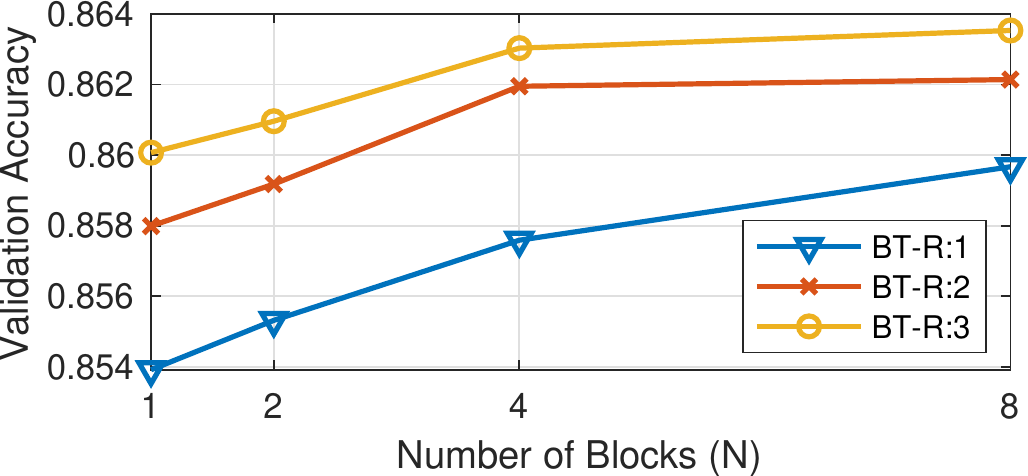}
	\caption{The test accuracy versus the BT-ranks on CIFAR10 dataset.} \label{e:fig:BTrank}
\end{figure}

\begin{figure}[t]
	\centering
	\includegraphics[width=0.45\textwidth]{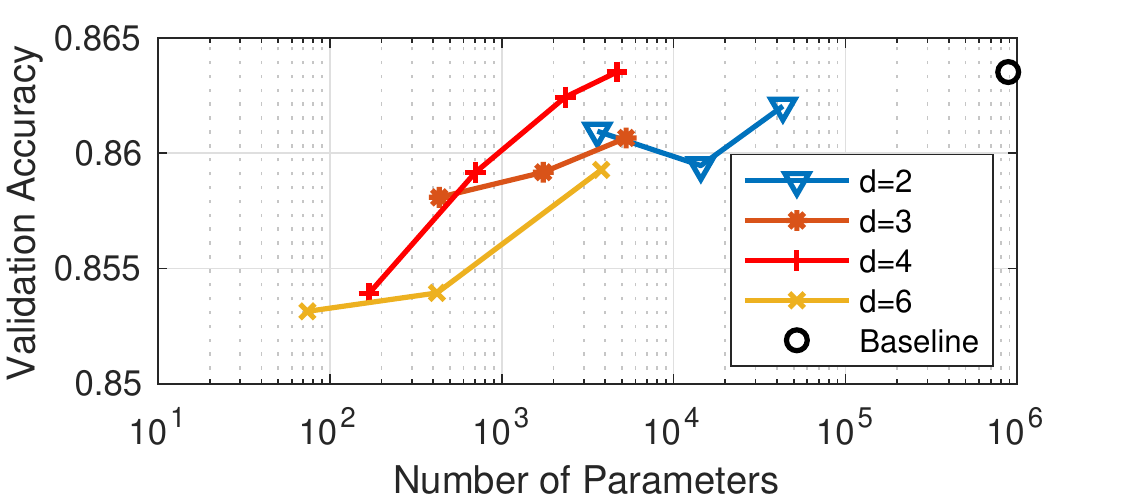}
	\caption{The test accuracy versus the number of cores $N$ on CIFAR10 dataset. The numbers in the parentheses denote the input and the output dimensions.} \label{e:fig:cores}
\end{figure}

As a BT-layer has two kinds of ranks, e.g., CP-rank $(N)$ and Tucker-rank $(R)$, and the core-order $d$, we utilize the CIFAR10 dataset as an example to study their impacts on the performance.
The network architectures are designed as follows: we just replace the first FC-layer with BT-layer where the CP-rank varies from 1 to 8, the Tucker-rank from 1 to 3 and the tensor-order $N$ from 2 to 6.

The results about BT-ranks are reported in Fig.~\ref{e:fig:BTrank}. We can intuitively see that  higher test accuracy can be achieved with  larger BT-ranks. In detail, we observe that when the number of blocks (CP-rank) is small, the accuracy curves rise quickly, but when the CP-rank becomes large, the curves are almost horizontal. Similar observations can be found in the Tucker-rank. 
Thus, if we want to get a better performance, we need to consider both and let them be appropriate values. The results about $N$ are reported in Fig.~\ref{e:fig:cores}.
In general,  higher accuracy can be obtained with a larger $N$. But when $N$ is too small, the accuracy does not go up because a smaller $N$ cannot sufficiently capture the weight's higher-order features. However, when $N$ is large, the performance is not dissatisfied as well due to the too small dimensions. Hence, we choose $N=4$ in this paper.

\subsubsection{Extension of adding BT-layers to ResNet}
It is  noted in \cite{zhang2017defense} that the FC layer can play an important role in transfering knowledge, although there are no FC-layers in some recent structures, such as ResNet~\cite{he2016deep}, due to the high computational complexity. Following \cite{zhang2017defense}, we test the performance of adding an FC-layer, a TT-layer and a BT-layer respectively. As a baseline, we use the ResNet-20 architecture\footnote{https://github.com/tornadomeet/ResNet}. The dimensions of FC-layer are set as $1024\times 1024$, same as \cite{zhang2017defense}, and we reshape them as $8\times 8\times 4\times 4$ and $8\times 8\times 4\times 4$ respectively in TT-layer and BT-layer. The results are illustrated in Table \ref{e:table:acc_on_resnet}. We can see that adding an FC-layer reduces the accuracy by about 1\%, while adding a BT-layer with moderate ranks (e.g., ``BT-N:4-R:2'') which only has additional 1344 parameters, could bring 0.4\% higher accuracy than the baseline. This sufficiently demonstrates that BT-layer could improve the network's performance, probably due to its ability of connecting high level features with the labels.

\subsubsection{Visualization}
\begin{figure}[htbp]
\centering

\subfigure[Original image]{
\begin{minipage}[t]{0.3\linewidth}
\centering
\includegraphics[width=1in]{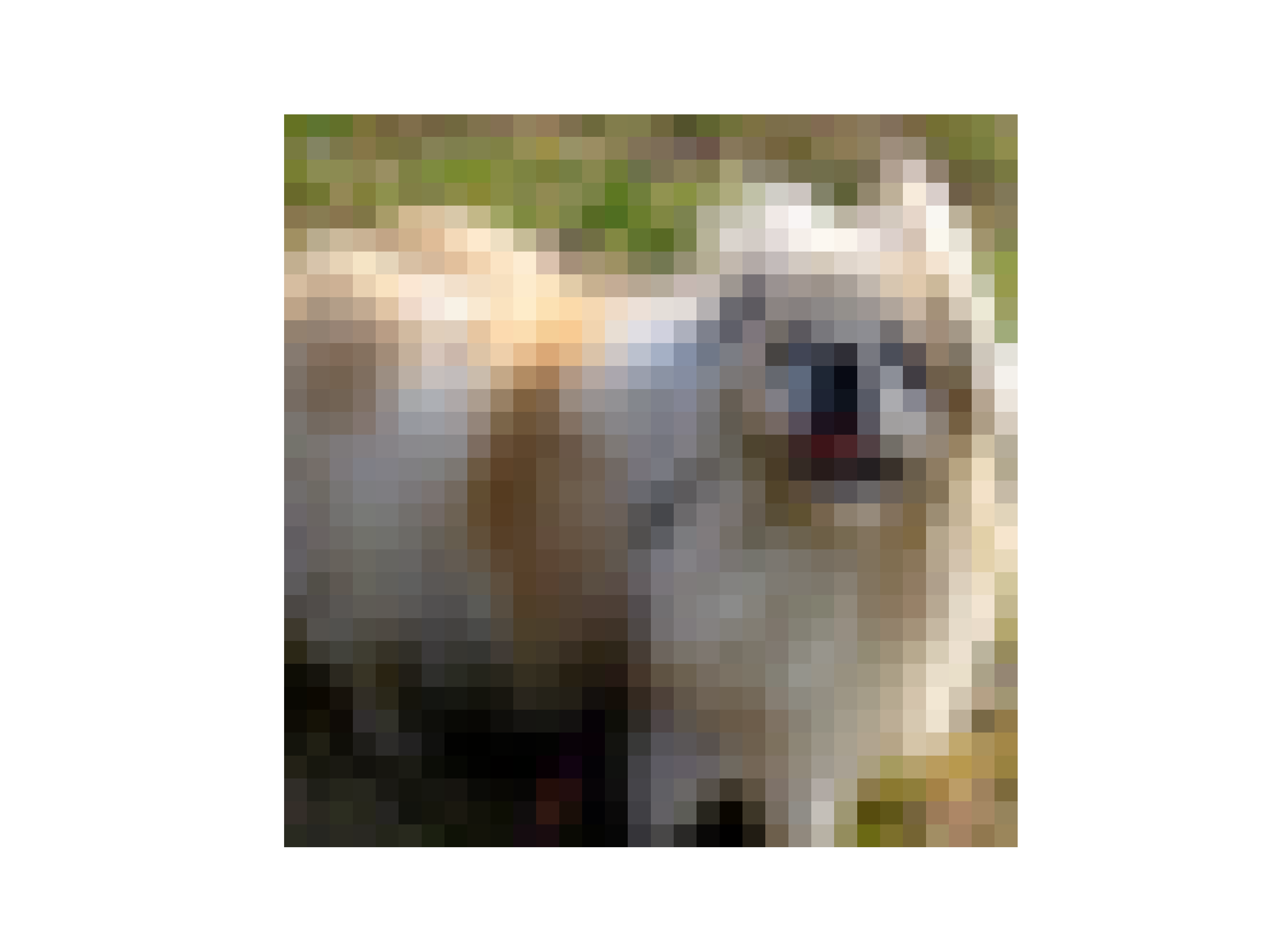}
\includegraphics[width=1in]{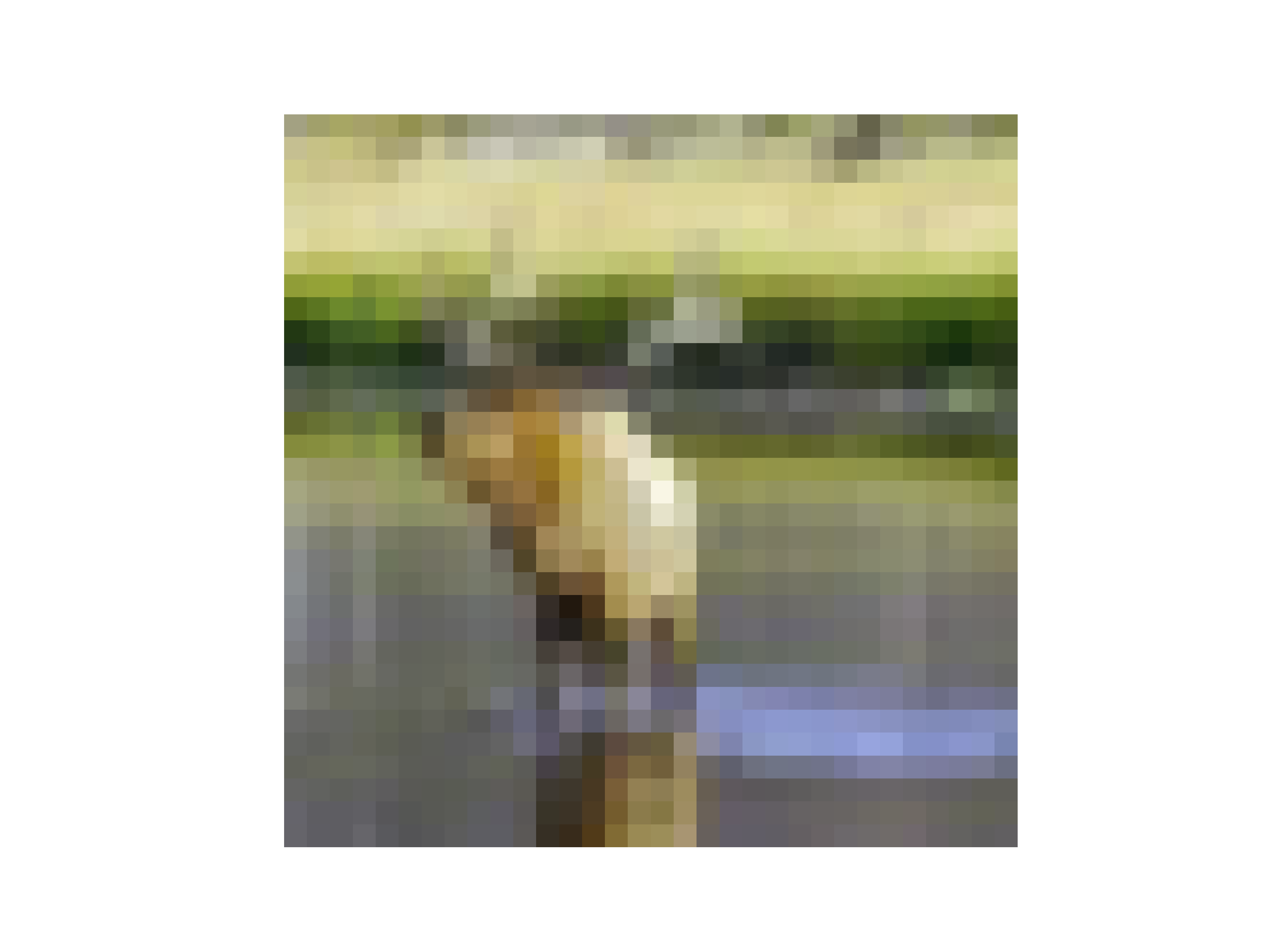}
\includegraphics[width=1in]{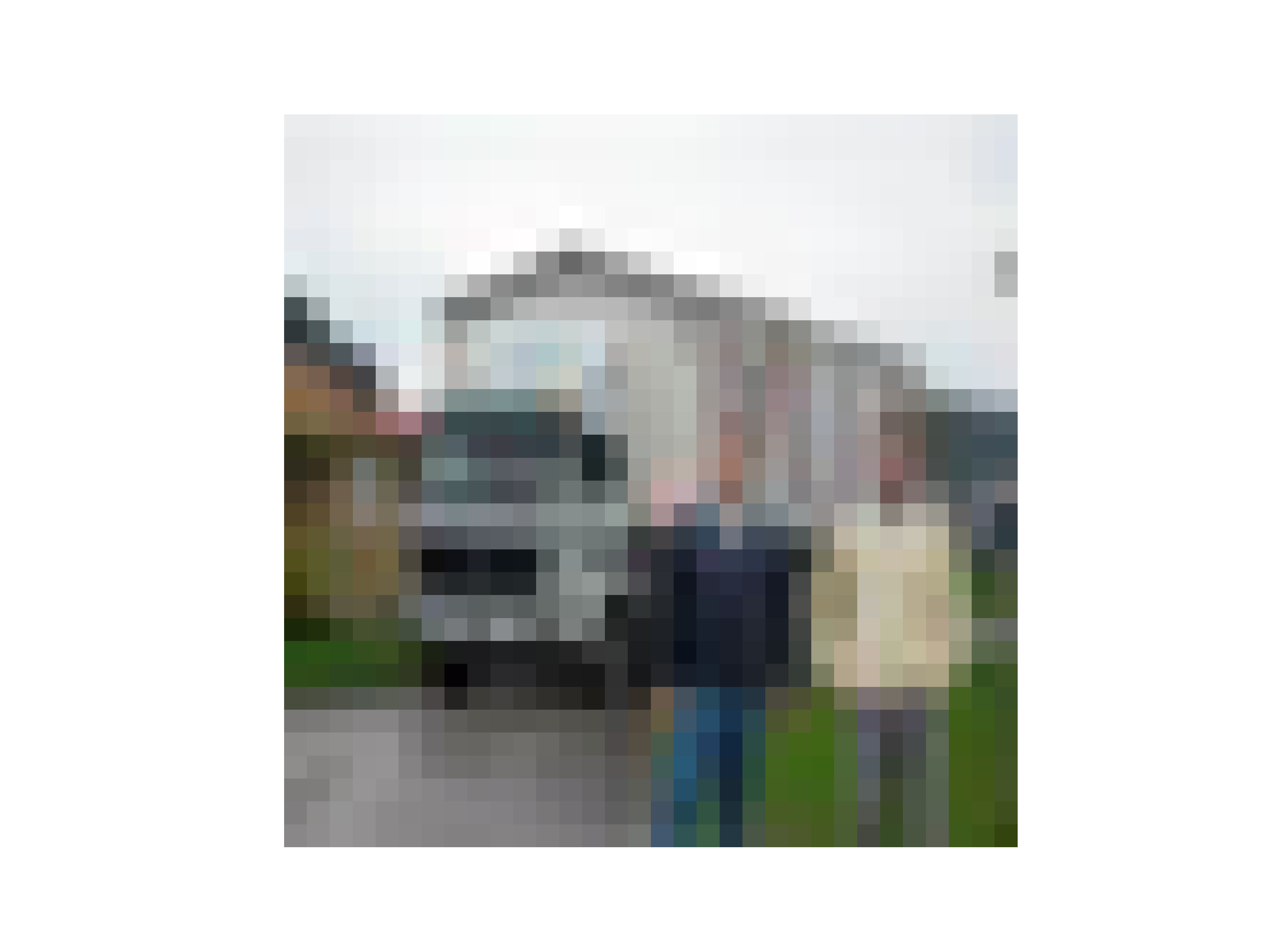}
\end{minipage}%
}%
\subfigure[Conv-layer]{
\begin{minipage}[t]{0.3\linewidth}
\centering
\includegraphics[width=1in]{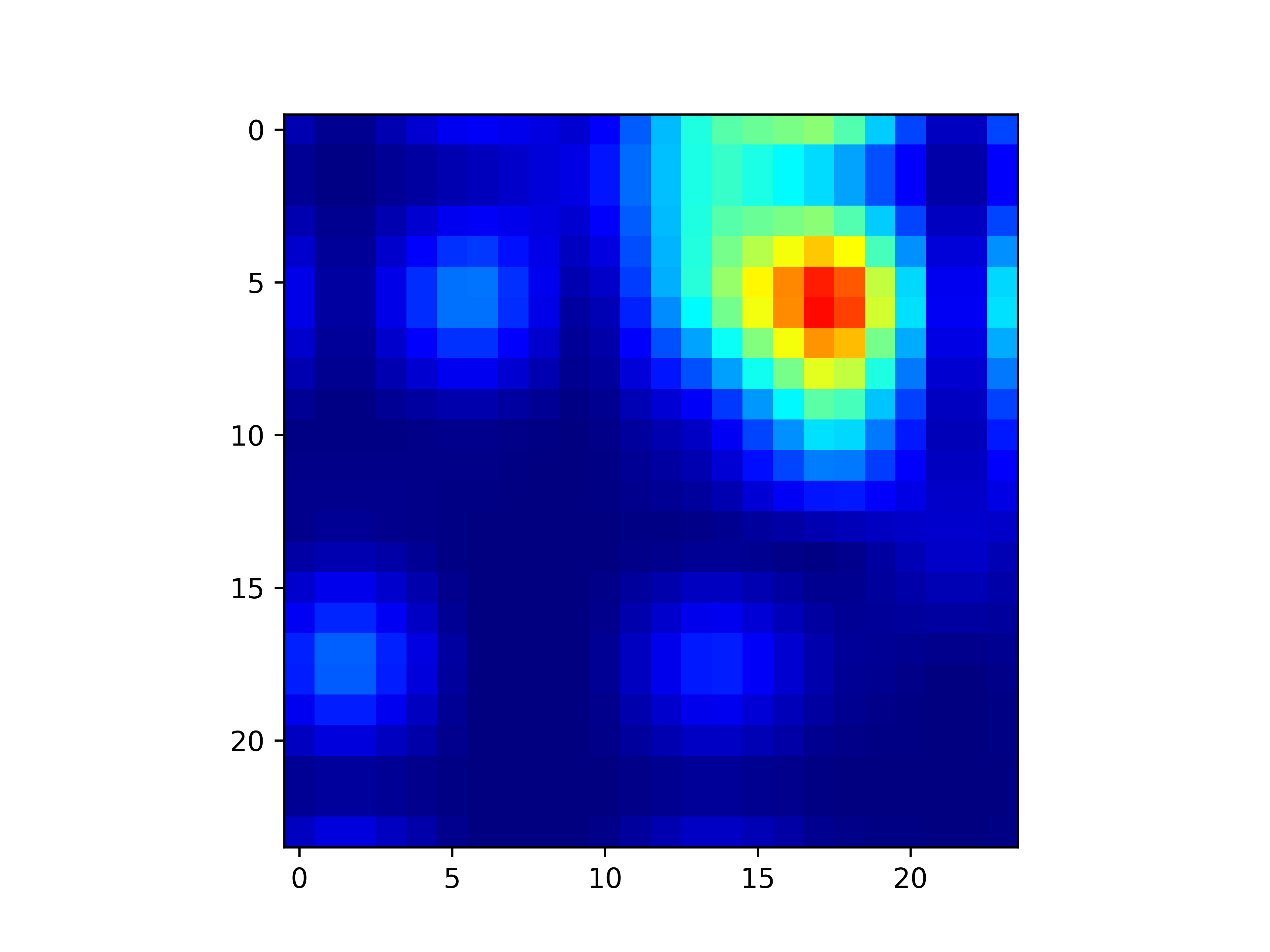}
\includegraphics[width=1in]{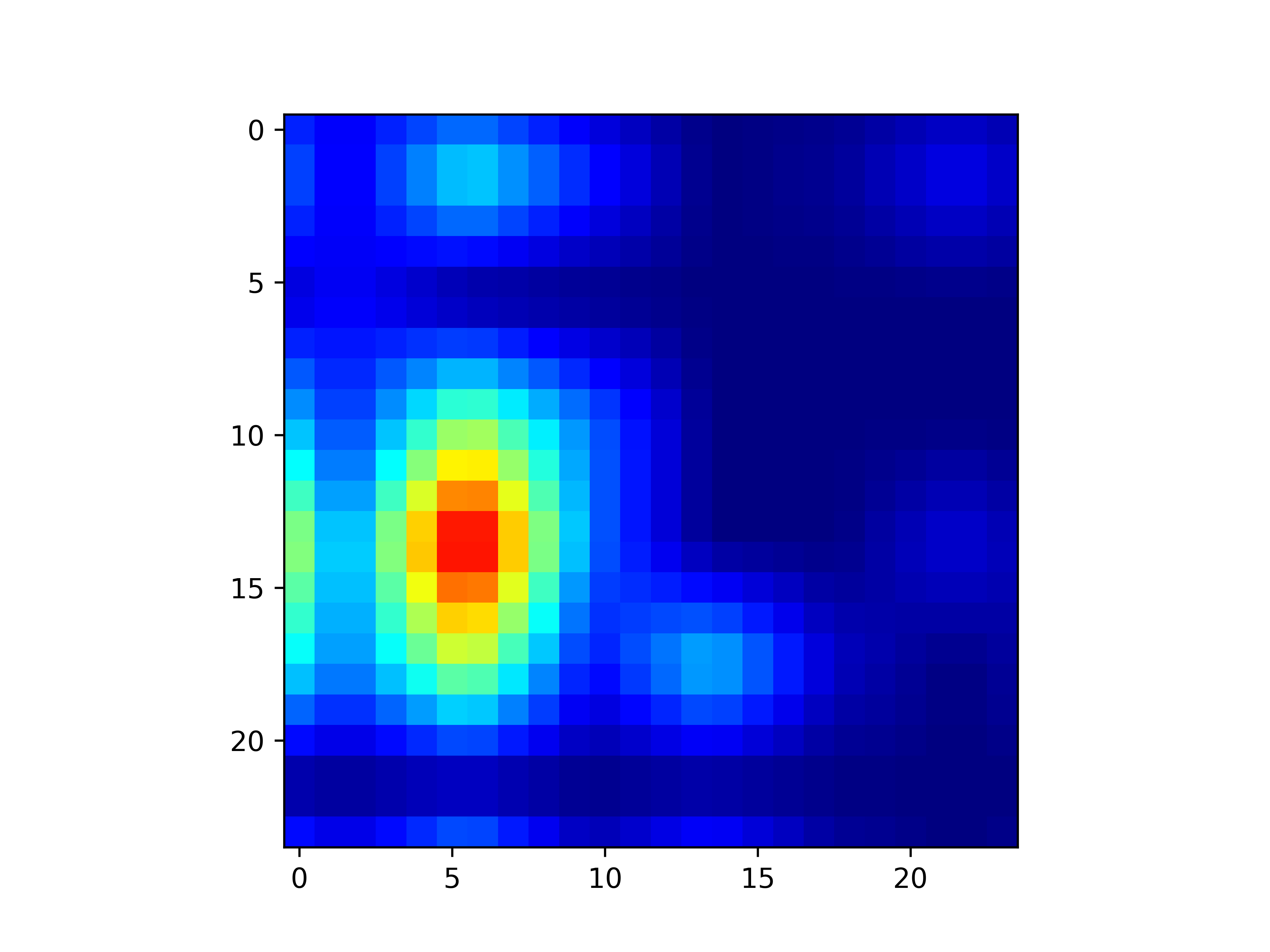}
\includegraphics[width=1in]{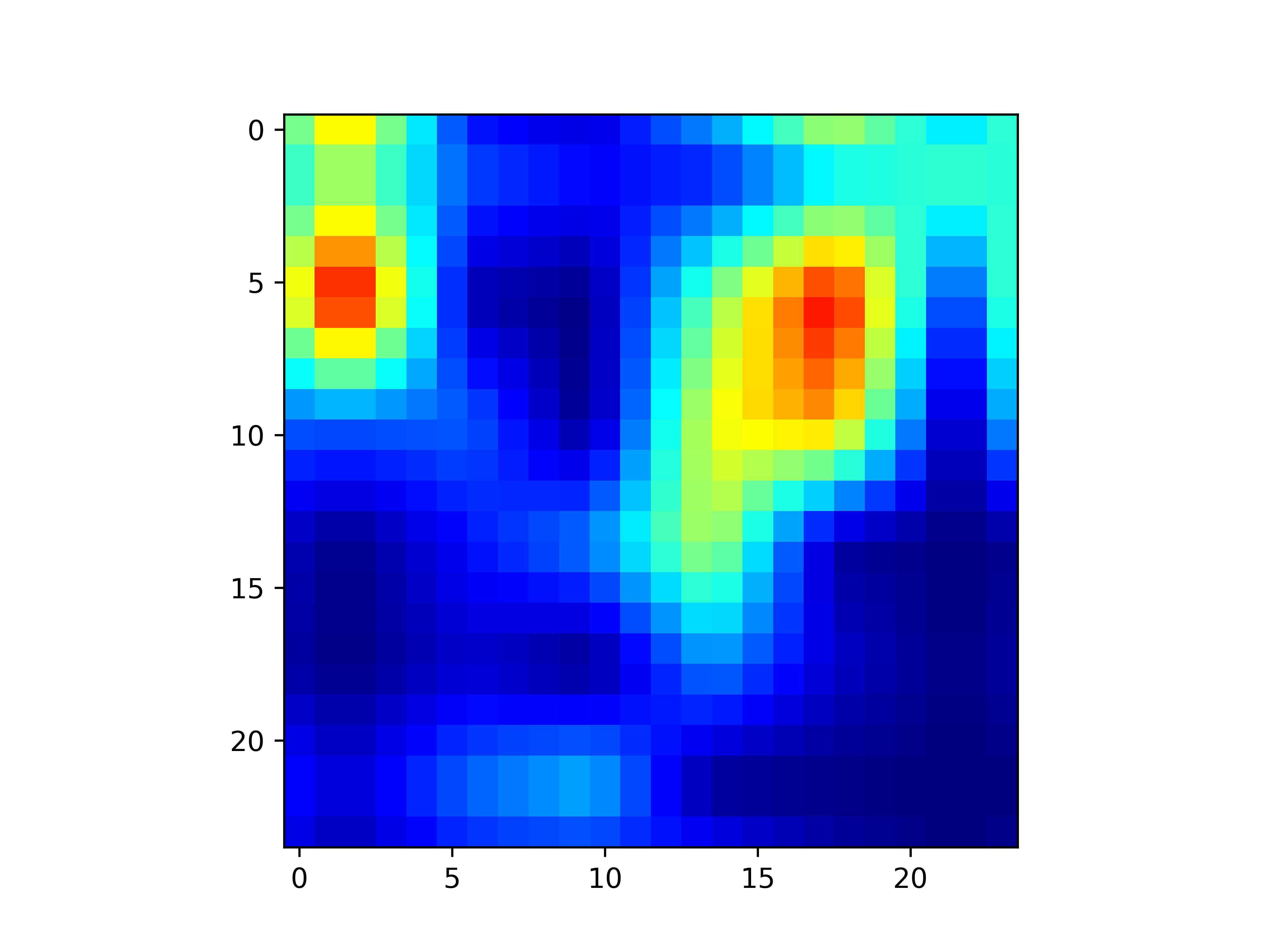}
\end{minipage}%
}%
\subfigure[BT-layer]{
\begin{minipage}[t]{0.3\linewidth}
\centering
\includegraphics[width=1in]{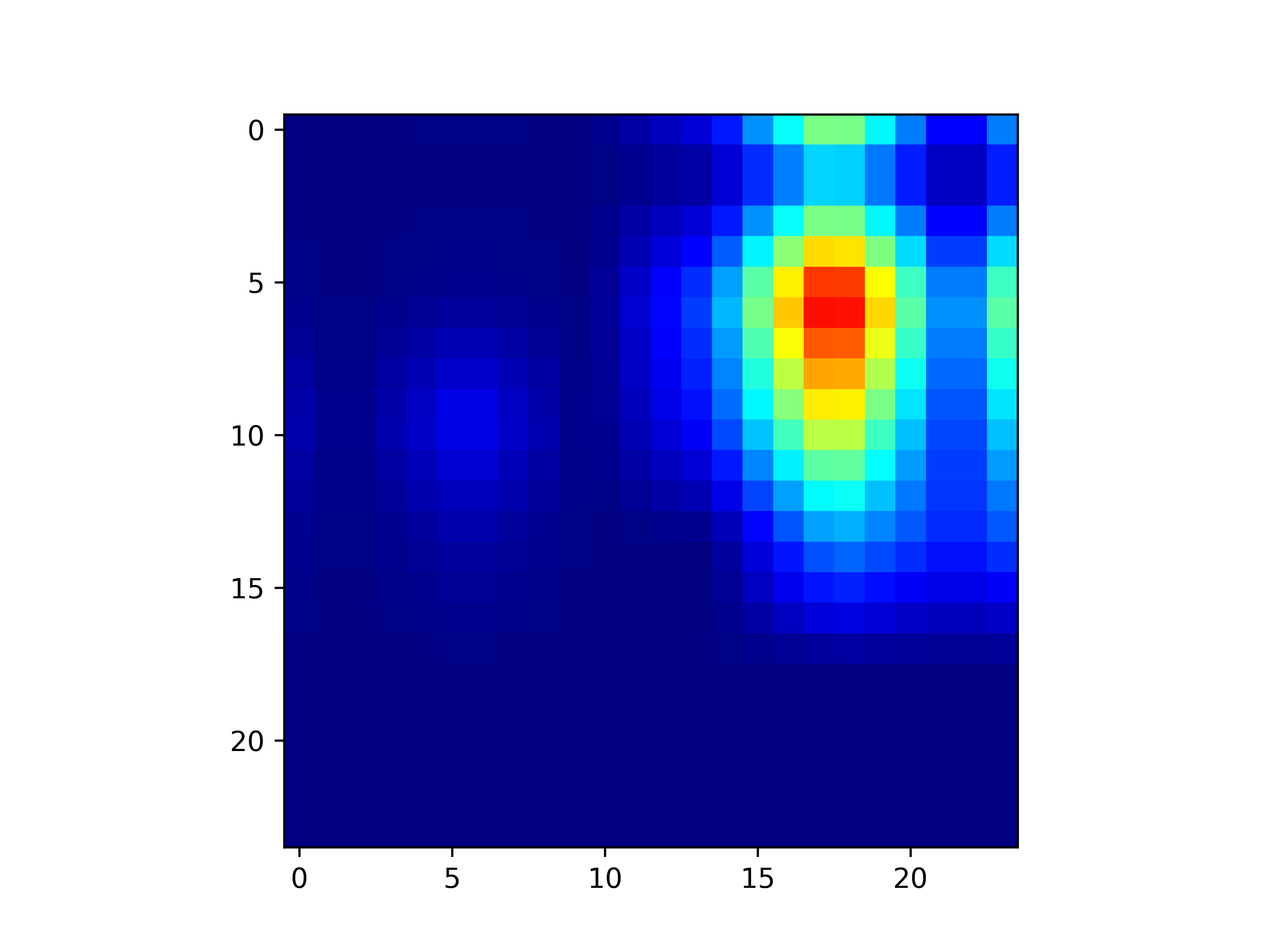}
\includegraphics[width=1in]{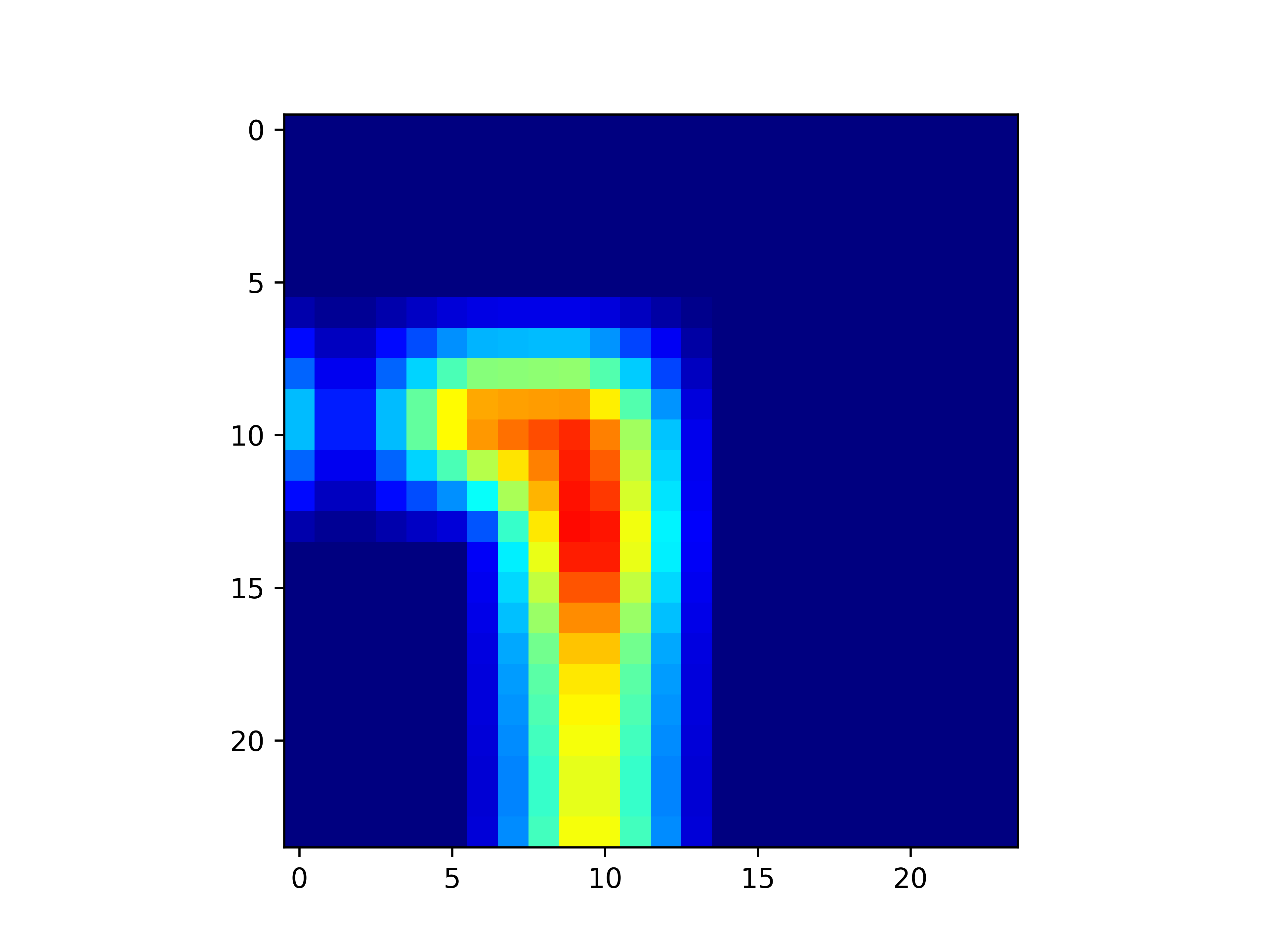}
\includegraphics[width=1in]{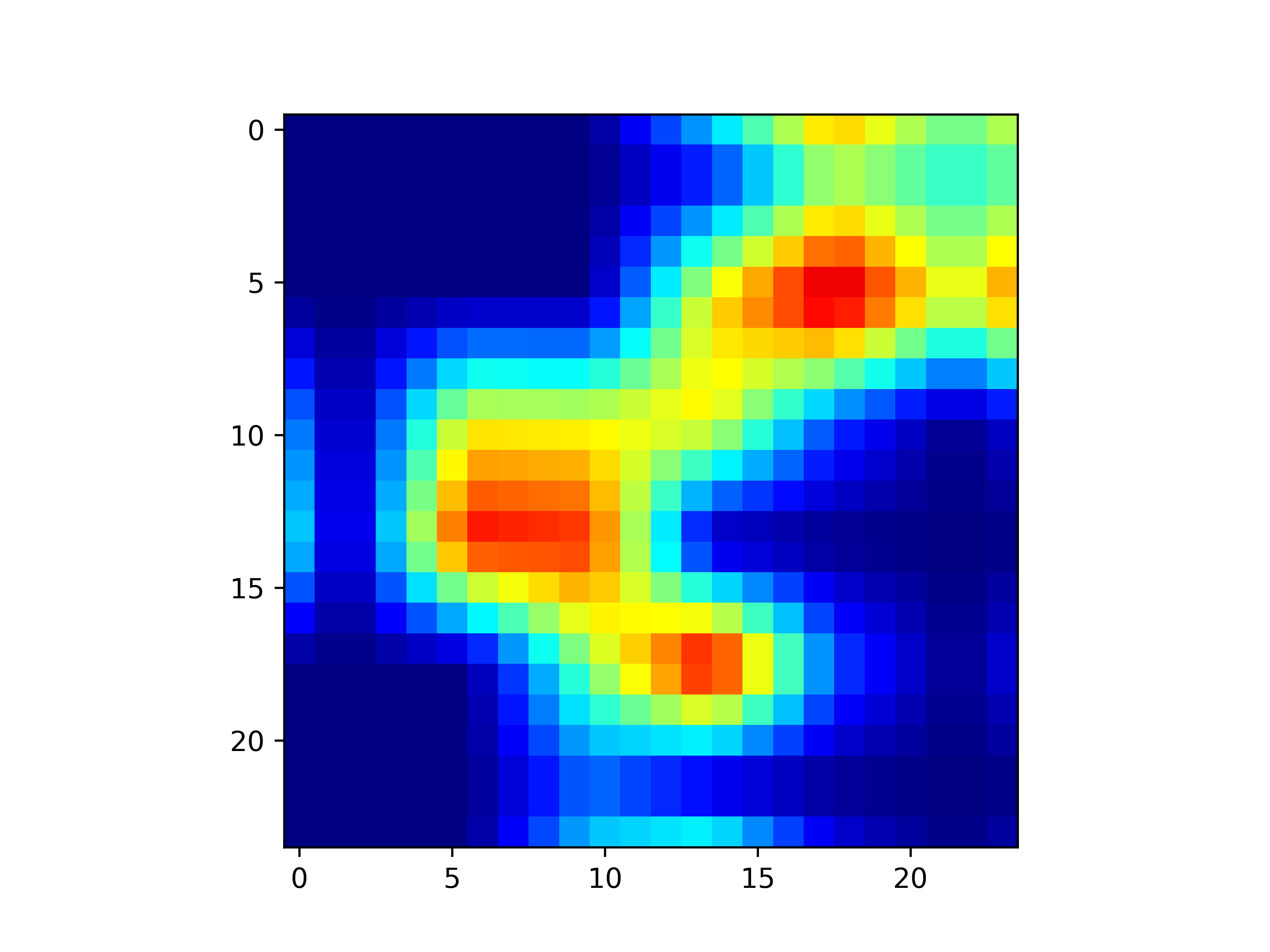}
\end{minipage}
}%

\centering
\caption{the result of visualization using Grad-CAM. (b) is the visualization of conv-layer and (c) is the visualization of BT-layer}
\label{fig:grad-CAM}
\end{figure}

For explaining and interpreting the proposed architectures, we apply the Grad-CAM~\cite{DBLP:journals/corr/SelvarajuDVCPB16} visualization techniques to the proposed architectures. Taking the CIFAR10 dataset as an example, we visualize the last conv-layer of CNN and then replace it with an BT-layer(here ``BT-N:2-R:3'' is used). The result of visualization is shown in Figure~\ref{fig:grad-CAM}. We can see from the figure that the BT-layer focuses more on key areas, e.g. the head of that dog here.


\subsubsection{Results of Running Time}
\begin{figure}[h]
	\centering
	\subfigure[Training time]{\label{e:fig:train_time}
		\includegraphics[width=0.43\textwidth]{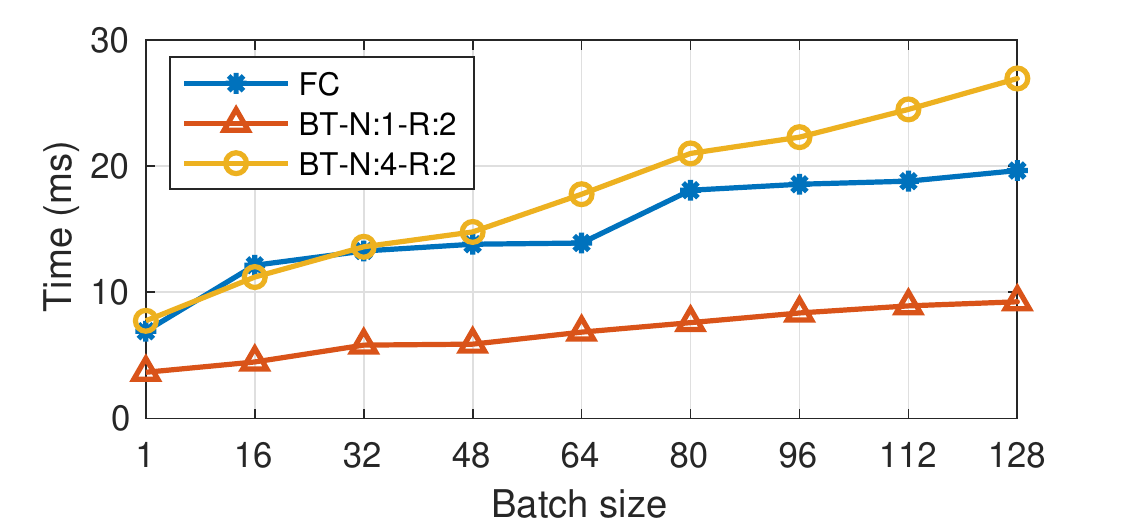}}
	\subfigure[Inference time]{\label{e:fig:inference_time}
		\includegraphics[width=0.43\textwidth]{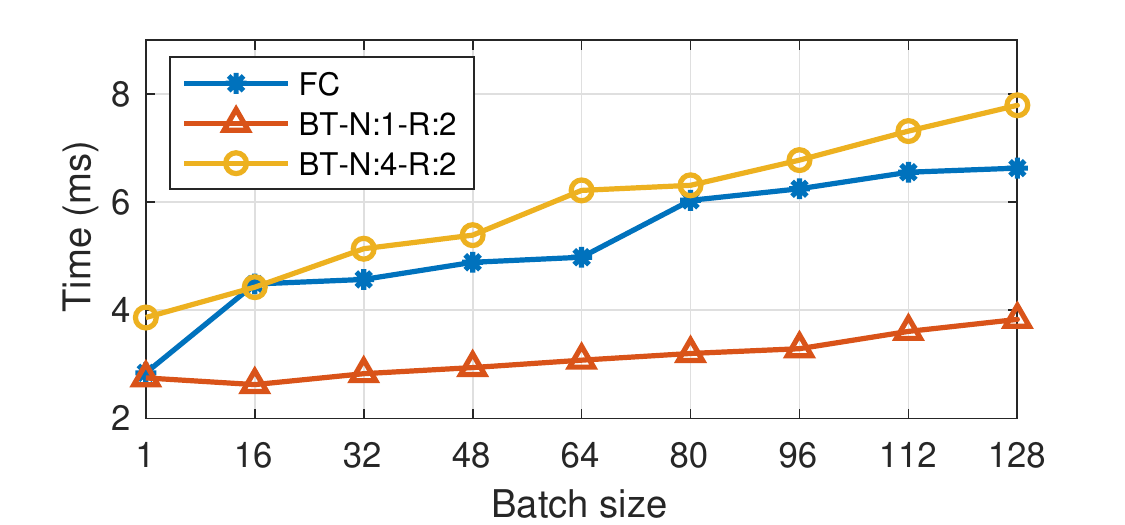}}
	\vspace{-2ex}
	\caption{Running time of $6400\times 4096$ FC-layer and its corresponding BT-layers.}
	\label{e:fig:running_time}
\end{figure}

In order to explore the time consumption of BT-layers, we test on a single $6400\times 4096$ FC-layer and the corresponding BT-layers which have the same architectures as in the ImageNet experiment.
These benchmark experiments are performed on TensorFlow with a single Tesla k40m GPU.
The results of the training time and inference time with different batch sizes are showed in Fig.~\ref{e:fig:running_time}.
From the results we can intuitively observe that when BT-ranks are small, BT-layer has a significant acceleration effect compared with FC-layer. When BT-ranks are increased to let the BT-Net catching the original performance, the time cost in BT-layer is also competitive with FC-layers.

\section{Evaluation on BT-RNN}
\label{sec:rnn}
We present our experiments conducted in two different parts:
1) quantitative evaluation. We use a video classification task show that compares to vanilla LSTM and TT-LSTM, our model achieve better performance while consuming fewer parameters, since BT model can capture more spatial information in fewer parameters.
2) qualitative evaluation. An image generation task and an image caption task will be conducted to show that BT model can handle various input data and output high-quality result.

\subsection{Evaluations of BT-LSTM on the Task of Video Classification}

\begin{figure}[t]
	\centering
	\subfigure[][Training loss of baseline LSTM, TT-LSTM and BT-LSTM.]{
		\includegraphics[width=0.45\textwidth]{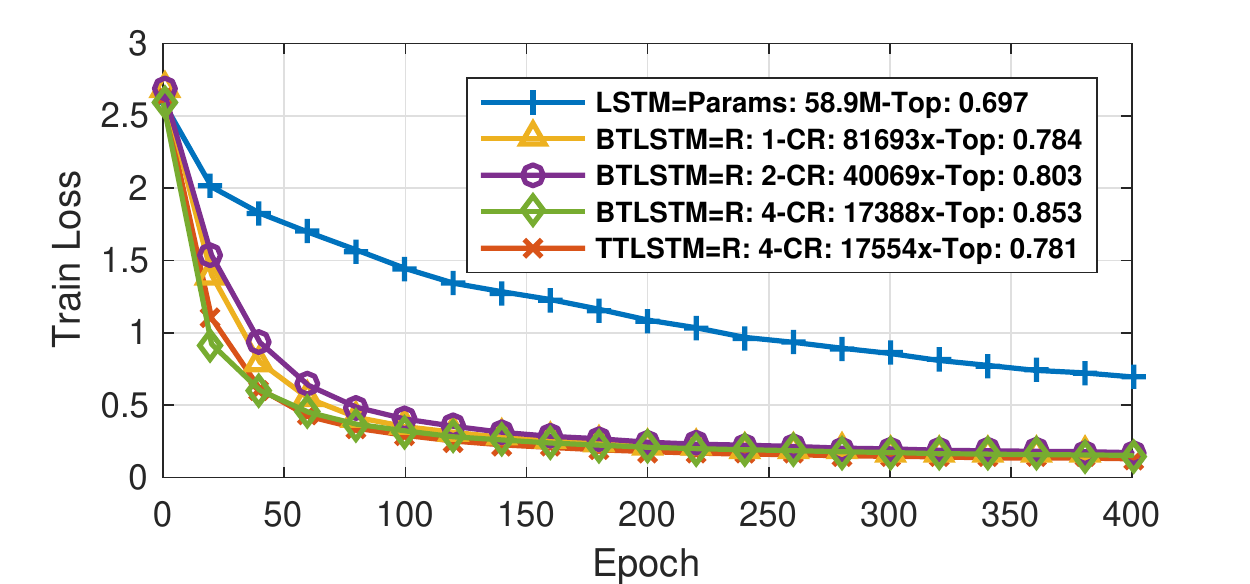} \label{fig:ucf11_loss}}

	\subfigure[][Validation Accuracy of baseline LSTM, TT-LSTM and BT-LSTM.]{
		\includegraphics[width=0.45\textwidth]{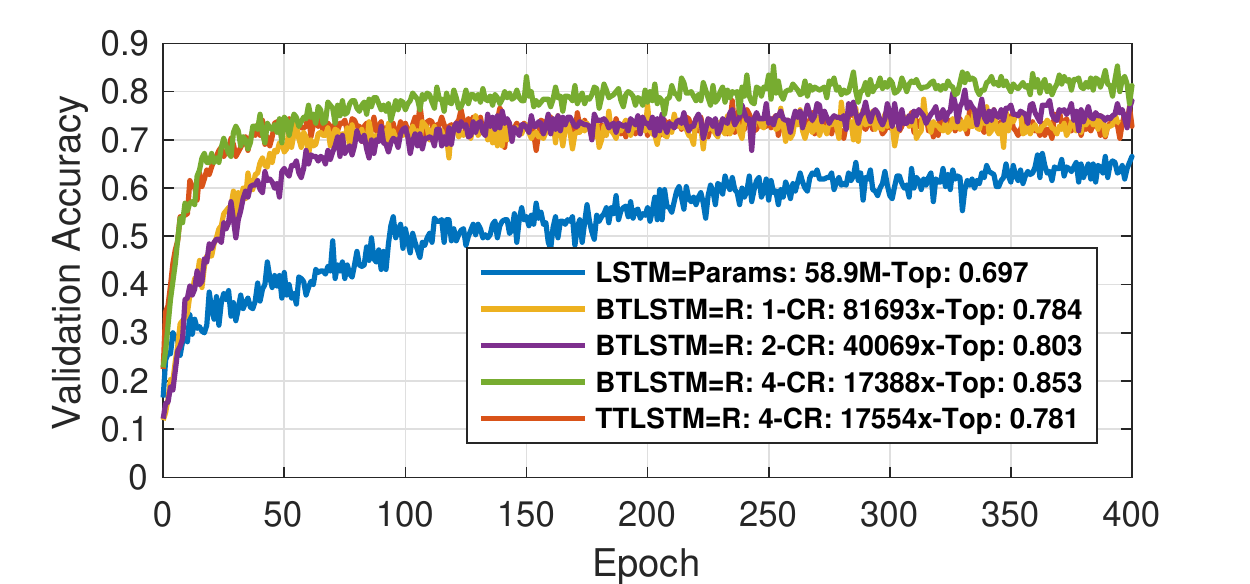} \label{fig:ucf11_acc}}
	\vspace{-1ex}
	\caption{Training and validation results of various RNN architectures. We use ``CR'' to denote the compression ratio.}
	\vspace{-2ex}
	\label{fig:loss_convergence_ucf11}
\end{figure}

\begin{table}[t]
	\small
	\begin{center}
		\begin{tabular}{c|c|c}
			\hline
			 & Method                                   & Accuracy       \\
			\hline\hline
			\multirow{3}{*}{\shortstack{Orthogonal                       \\ Approaches}} & Original \cite{liu2009recognizing} & 0.712 \\
			 & Spatial-temporal \cite{liu2013spatial}   & 0.761          \\
			 & Visual Attention \cite{sharma2015action} & 0.850          \\
			\hline
			\multirow{3}{*}{\shortstack{RNN                              \\ Approaches}}
			 & LSTM                                     & {0.697}        \\
			 & TT-LSTM \cite{yang2017tensor}            & 0.796          \\
			 & BT-LSTM                                  & \textbf{0.853} \\
			\hline
		\end{tabular}
	\end{center}
	\caption{Some important results reported in literature on the UCF11 dataset with the video classification task.}
	\vspace{-2ex}
	\label{tbl:ucf11_acc_state_of_the_art}
\end{table}

\begin{figure*}[t]
	\captionsetup[subfigure]{labelformat=empty}
	\centering
	\subfigure[][\scriptsize{ \textbf{Truth}: A sleepy cat snuggling with a stuffed animal. \newline
      \textbf{BT-LSTM}: A cat laying on a bed with a stuffed animal.\newline
      \textbf{LSTM}: A dog and a dog are sleeping on a bed.
    }]{
    \includegraphics[width=0.20\textwidth, height=0.25\columnwidth]{pics/mscoco/COCO_val2014_000000489743.jpg}} \quad
  \subfigure[][\scriptsize{ \textbf{Truth}: A person on a surfboard in the water. \newline
      \textbf{BT-LSTM}: A man on a surfboard riding a wave.\newline
      \textbf{LSTM}: A man riding a wave on top of a surfboard.
    }]{
    \includegraphics[width=0.20\textwidth, height=0.25\columnwidth]{pics/mscoco/COCO_val2014_000000210915.jpg}} \quad
  \subfigure[][\scriptsize{ \textbf{Truth}: Three people riding horses on the beach right up to the water. \newline
      \textbf{BT-LSTM}: A group of people riding horses on the beach.\newline
      \textbf{LSTM}: A group of people standing on a beach with surfboards.
    }]{
    \includegraphics[width=0.20\textwidth, height=0.25\columnwidth]{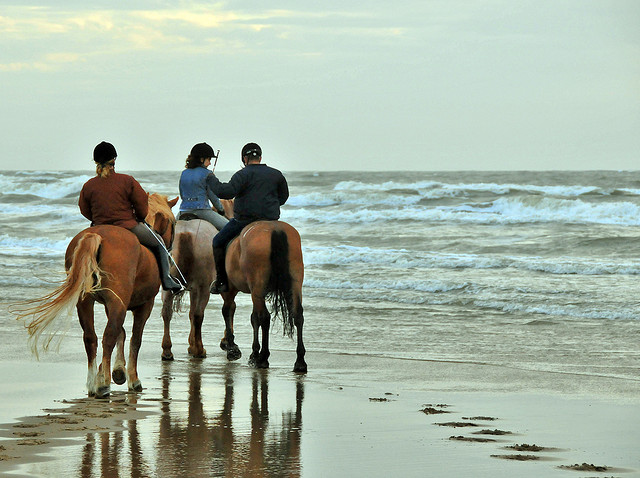}} \quad
  \subfigure[][\scriptsize{ \textbf{Truth}: This is a group of elephants on a grassy plane.  \newline
      \textbf{BT-LSTM}: A herd of elephants walking across a grassy field.\newline
      \textbf{LSTM}: A herd of elephants walking in a grassy area.
    }]{
    \includegraphics[width=0.20\textwidth, height=0.25\columnwidth]{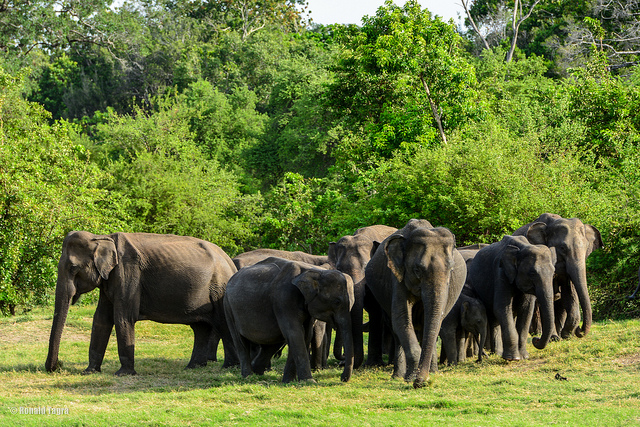}}
    \caption{Demonstrations of image captioning in MSCOCO dataset. We can see that BT-LSTM has little improvement than the captions generated by LSTM.}
	\label{fig:mscoco}
\end{figure*}





UCF11 YouTube Action \cite{liu2009recognizing} is a large-scale dataset used in video classification task, containing 1600 video clips with 11 action categories. A video clip can be seen as a sequence of image frames naturally. But it's extremely difficult to train an RNN model to accomplish this task as the high dimension input data. Our data preprocessing follows the procedure reported by \cite{yang2017tensor}, which down-sampling from $320 \times 240 \times 3$ to $160 \times 120 \times 3$ and 6 frames will be random sampled for each video clip. The frames will be fed into LSTM and BT-LSTM model in sequential setup.

To evaluate the performance of BT-LSTM against LSTM and TT-LSTM, we construct the classification model with a single RNN cell with various types. Fig.~\ref{fig:loss_convergence_ucf11} demonstrates the performance of 3 architectures with various hyper-parameter settings. Since the RNN design philosophy is kept in our model, the other orthogonal approaches such as  \cite{sharma2015action} and \cite{liu2013spatial}, will be equipped together to improve the accuracy.

The hyper-parameters of three architectures are set as follows: all the hidden size in RNN blocks are set as 256; all the Core-order of BT-LSTM models are set as $d=4$; the input data are tensorized as $I_1 = 8, I_2 = 20, I_3 = 20, I_4 = 18$; while the rank of TT-LSTM is $R_1 = R_5 = 1, R_2 = R_3 = R_4 = 4$ and the Tucker-ranks in BT-LSTM models are various.

The training and validation results of BT-LSTM and the comparison methods are presented in Fig.~\ref{fig:loss_convergence_ucf11}. The number of parameters of the vanilla LSTM model is 58.9M, while the BT-LSTM with Tucker-rank 1,2,4, is 721, 1470, and 3387 respectively, raising a very high compression ratio. However, the BT-LSTM models still outperform the vanilla LSTM model with the highest 15.6\%, saying that the BT-LSTM model is an effective way to reduce the model parameter while improving model accuracy. We also reproduce the TT-LSTM and find that with comparable parameters, BT-LSTM(R=4) outperforms TT-LSTM(R=4) with 7.2\%.

With much less parameters, BT-LSTM raises a higher convergence speed than the vanilla LSTM. According to Fig.~\ref{fig:ucf11_acc}, BT-LSTM achieves the validation accuracy of 60\% at the 16th epoch, while the vanilla LSTM needs about 220 epochs. TT-LSTM has nearly the same parameters with BT-LSTM, so as the convergence speed.

Table. \ref{tbl:ucf11_acc_state_of_the_art} gives the state-of-the-art results in UCF11 dataset. Comparing to other methods, our BT-LSTM provides a more elegant solution with sparse connection in LSTM architecture.

\subsection{Evaluations of BT-LSTM on Image Captioning Task}
Image Captioning is a challenging task that generates a readable sentence to describe an image. The base model we used is the Neural Image Captioning\cite{vinyals2015show} and the dataset is the large-scale MSCOCO\cite{lin2014microsoft} dataset. MSCOCO contains 82783 images for training and 40775 images for testing. The model uses a pre-trained Inception-V3 model as the backbone architecture to extract the image features and a sentence-to-sentence LSTM to generate sentence. We also follow the data pre-processing described in \cite{vinyals2015show} that scales images to $224 \times 224$ and subtracts the channel means.

In this model, the input data of BT model is heterogeneous vectors merged by the image feature vector generated by CNN backbone and word word feature vectors generated by embedding layers.
According to Fig.~\ref{fig:mscoco}, both LSTM and BT-LSTM can pronounce human-readable sentences, but the sentences generated by BT-LSTM is more close to the ground truth.

\section{Related Work}
\label{sec:related}
Extensive research on DNNs has proven that the dense connection is significantly inefficient at extracting the spatially latent local structures and local correlations naturally exhibited in the data~\cite{lecun1995convolutional, han2015learning}. 

Reducing the computation or storage complexity  is the holy grail of the neural networks design. There have been a number of network slimming techniques motivated from different perspectives~\cite{cheng2017survey}, including network hashing~\cite{chen2015compressing}, network sparsification or pruning~\cite{srivastava2014dropout,HanMD15,narang2017exploring,DBLP:journals/corr/MolchanovTKAK16,narang2017block,Wen2017LearningIS}, binarization and quantization~\cite{Hubara2016BNN,Gong2014CompressingDC}, low rank representation~\cite{Sironi15filter,Denton2014ELS,Zhang2016AVD,jaderberg2014speeding,tjandra2017compressing,yu2017long,Ye_2018_CVPR,WangSEWA18,WangZPXX19,PanXWYWBX19}, and knowledge distillation~\cite{hinton2015distilling}. 

Among these methods, the low-rank methods is a  type of attractive approach to implicitly prune connections, which employs matrix factorization or tensor factorization techniques~\cite{XuYQ12icmltensor,ChenLKX13,orus2014practical,XuYQ15tensor,ZheZWLXQG16,LiuHLZX18,HaoLYX18} on the weighting matrices to obtain compact weight matrices or tensors. 
Some of the traditional tensor decomposition methods have been applied in CNNs or RNNs to alleviate the inefficient fully-connected structure~\cite{novikov2015tensorizing, yang2017tensor, bai2017tensorial, tjandra2017compressing, kossaifi2017tensor}, achieving higher accuracy or better parameter reduction than the traditional matrix decomposition based methods\cite{sainath2013low,xue2013restructuring,schollwock2011density}.
The accuracy improvement is mainly from the capture of the high-order spatial information. In addition, tensor decomposition can also be used in accelerating the convolution operation.

Recently, tensor networks ~\cite{osti_22403404,Cichocki2016TND,tensornetwork19} has attracted the attention of researchers in the machine learning area~\cite{glasser2018supervised,han2018unsupervised,stoudenmire2016supervised}. Recently, Novikov et al. applied the Tensor Train method to replace the fully-connected layer in CNNs, which can reduce a huge amount of parameters while speeding up the inference phase~\cite{novikov2015tensorizing}. Tensor Train based structure also applied in RNNs to solve the high dimensional input data~\cite{yang2017tensor}. Other tensor network decomposition methods also applied in Deep Neural Networks (DNNs) for various purposes~\cite{lebedev2014speeding, yunpeng2017sharing, kossaifi2017tensor,Wang2018WideCT,2018arXiv180510352S,bttensor_attention19}.

It is important to note that although recent leading CNNs architectures, e.g., DenseNet \cite{huang2016densely} and ResNet \cite{he2016deep}, try to cut out the huge fully connected layer, there are no theories to prove the necessity of avoiding fully connected layers. In addition, it has been shown that the fully connected layers have good knowledge transfer ability~\cite{zhang2017defense}. Furthermore, we show that a block-term tensor layer can be added into ResNet to further improve the performance, as demonstrated in Table~\ref{e:table:acc_on_resnet}.

\section{Conclusion}
\label{sec:con}

We have presented a new network architecture design in which the commonly used fully-connected layers in CNNs and RNNs are replaced with the Block Term-layers. 
Thanks to the better representation ability of the block-term tucker structure, BT-layers can not only greatly reduce the number of parameter size, but also improve the generalization performance especially for recurrent neural networks. 
Our experiments on several datasets have demonstrated the promising performance of the BT-layers in both CNNs and RNNs.


As for future work, in order to obtain higher compression ratio for real applications, we plan to combine the BT-Nets with other compression techniques, such as pruning and binarization.

\section*{Acknowledgement}
This paper was partially 
Supported by the National Key Research and Development Program of China (No. 2018AAA0100204). Dr. Haiqin yang is also affiliated with Dept. of Decision Sciences and Managerial Economics, The Chinese University of Hong Kong, Hong Kong, and Dept. of Computing, The Hang Seng University of Hong Kong, Hong Kong. 
We thank Dr. Shuicheng Yan and Mr. Linnan Wang for valuable comments and discussions to this paper, and thank Dr. Xinqi Chu for providing computing resources.


\bibliography{mybibfile}
\bibliographystyle{elsarticle-num}
\end{document}